\documentclass[10pt,twocolumn,letterpaper]{article}

\usepackage[arxiv]{cvpr}         

\usepackage{booktabs}
\usepackage{multirow}
\usepackage{adjustbox} 
\usepackage{placeins}
\makeatletter
\newcommand{\suppfloatsetup}{%
    \let\supp@table\table \let\supp@endtable\endtable
    \let\supp@figure\figure \let\supp@endfigure\endfigure
    \renewenvironment{table}[1][]{\supp@table[!hbp]}{\supp@endtable}%
    \renewenvironment{table*}[1][]{\supp@table[!hbp]}{\supp@endtable}%
    \renewenvironment{figure}[1][]{\supp@figure[!hbp]}{\supp@endfigure}%
    \renewenvironment{figure*}[1][]{\supp@figure[!hbp]}{\supp@endfigure}%
    \renewcommand{\topfraction}{0.95}\renewcommand{\bottomfraction}{0.95}%
    \renewcommand{\textfraction}{0.05}\renewcommand{\floatpagefraction}{0.5}%
    \setcounter{topnumber}{4}\setcounter{bottomnumber}{4}\setcounter{totalnumber}{6}%
}
\makeatother
\usepackage[resetlabels]{multibib}
\makeatletter
\providecommand*{\@extra@binfo}{}\providecommand*{\@extra@b@citeb}{}
\newcites{S}{References} 
\let\mb@citeS\citeS \protected\def\citeS{\mb@citeS} 
\AtBeginDocument{\immediate\write\@mainaux{\gdef\string\@extra@binfo{}}}
\newcommand{\suppbibsetup}{%
    \gdef\@extra@b@citeb{@S}
    \renewcommand{\citenumfont}[1]{S##1}%
    \renewcommand{\bibnumfmt}[1]{[S##1]}%
    \@ifundefined{Hy@backout}{}{%
        \let\mb@orig@backout\Hy@backout
        \def\Hy@backout##1{{\let\@auxout\@mainaux\mb@orig@backout{##1}}}}%
}
\renewcommand{\bibliographyS}[1]{%
    \if@filesw
        \immediate\write\@auxoutS{\string\bibdata{#1}}%
        \immediate\write\@mainaux{\gdef\string\@extra@binfo{@S}}%
    \fi
    \begingroup
        \let\refname\refnameS \let\bibname\refnameS
        \@input@{S.bbl}%
    \endgroup
}
\makeatother
\AtEndPreamble{
    \crefname{suppsection}{Supp.~Sec.}{Supp.~Secs.}
    \Crefname{suppsection}{Supp.~Section}{Supp.~Sections}
    \crefname{supptable}{Supp.~Tab.}{Supp.~Tabs.}
    \Crefname{supptable}{Supp.~Table}{Supp.~Tables}
    \crefname{suppfigure}{Supp.~Fig.}{Supp.~Figs.}
    \Crefname{suppfigure}{Supp.~Figure}{Supp.~Figures}
    \crefname{suppequation}{Supp.~Eq.}{Supp.~Eqs.}
    \Crefname{suppequation}{Supp.~Equation}{Supp.~Equations}
}

\definecolor{cvprblue}{rgb}{0.21,0.49,0.74}
\usepackage[pagebackref,breaklinks,colorlinks,allcolors=cvprblue]{hyperref}

\def\confName{CVPR}
\def\confYear{2027}

\title{One Subspace Is Not Enough: LocalSubspaceAD for Training-Free Few-Shot Anomaly Detection }

\author{Daniel Pröll$^{1,2}$
\qquad Sebastian Hegenbart$^{2,3}$\\
{\small $^1$University of Salzburg, Austria \quad
$^2$Digital Factory Vorarlberg GmbH, Austria \quad
$^3$University of Applied Sciences Vorarlberg, Austria}\\
{\tt\small \{daniel.proell, sebastian.hegenbart\}@vactory.at}
}

\begin{document}
\maketitle
\begin{abstract}
Defect-free industrial samples can vary considerably in appearance, texture, and geometry. In few-shot anomaly detection, this variability has to be modeled from only a handful of images. SubspaceAD showed that a single principal subspace of foundation-model patch features is a strong training-free model of nominal appearance. However, a single subspace has to cover all of the nominal data's variability, so a deviation that is anomalous for one semantic region can be within the nominal variability of another. We propose LocalSubspaceAD, which extends SubspaceAD from a single subspace to a mixture of subspaces, assigning each group of patches with similar features its own subspace.
The set of patch-features is partitioned using k-means, one subspace is fit per cluster, and each patch is scored by the minimum of re-projection residuals over all subspaces. 
The proposed method requires no training and no EM refinement. On MVTec AD, VisA and Real-IAD with 1, 2 and 4 shots, it improves over a single subspace (SubspaceAD) on both evaluated backbones: by $1.2$ to $1.8$ points image AUROC on DINOv2, and by $1.4$ to $4.4$ points in image AUROC, pixel AUPR and AUPRO on DINOv3. The gains are most pronounced on objects with heterogeneous nominal appearance, such as multi-instance categories and PCBs. Ablations show that the gains stem from a finer-grained model of variability through locally oriented subspaces, rather than from the subspace offsets (cluster means). We further show that the gains do not stem from added capacity alone: the local subspaces appear to capture the structure of coherent clusters in feature space, and the model does not merely act as a compressed memory bank.

\end{abstract}
 
\section{Introduction}
\label{sec:intro}

Nominal data in industrial inspection can be highly heterogeneous, particularly for objects with multiple instances of a part or complex assemblies such as printed circuit boards.

A model of the nominal state is required to capture all types of nominal variation in an image, including differences between semantic regions of an object and differences among multiple instances of the same part. To detect anomalies reliably, these variations need to be modeled tightly: the more variation the model accepts everywhere, the more room is left for anomalies to go undetected.

Foundation models provide patch embeddings that are strong enough for conceptually simple models of the nominal state, such as memory banks of features \cite{dammAnomalyDINOBoostingPatchbased2025,rothTotalRecallIndustrial2022,tangDuoADLeveragingCLS2026} and methods that combine textual descriptions with images \cite{jeongWinCLIPZeroFewShot2023,zhouAnomalyCLIPObjectagnosticPrompt2023}. Recently, SubspaceAD \cite{lenderingSubspaceADTrainingFreeFewShot2026} has shown that a single subspace fitted to these embeddings with principal component analysis (PCA) gives very good few-shot performance, scoring each patch by its residual to the subspace. A single subspace, however, has to cover all of the nominal data's variability. Rare nominal patterns carry little variance, which forces a trade-off: exclude them from the global model and similar patches become false positives; include them and they are accepted in every semantic region, potentially causing anomalies to be undetected. Moreover, different semantic regions of an object can exhibit different, even conflicting, variations that cannot be captured by a single model.

We therefore model the nominal patch embeddings as a mixture of local subspaces, which gives each semantic region its own subspace. LocalSubspaceAD partitions the support patches with k-means, fits one PCA per cluster and scores a test patch by its residual to the subspace that reconstructs it best. Like SubspaceAD, the method is training-free and purely statistical, and it reduces to SubspaceAD for a single component. We evaluate it with two backbones: DINOv2, with features averaged over layers as in SubspaceAD, and DINOv3, with one set of local subspaces per layer and the residuals normalized by a noise variance pooled over all components.

Modeling the nominal data with a mixture of local subspaces shows consistent improvement over a single global subspace. Ablation experiments examine where the improvement comes from. We separate the contribution of the local subspaces from that of the cluster means, replace the k-means clusters with random assignments of patches to test whether added capacity alone explains the gain, and compare against scoring by the distance to the cluster centers, as a compressed memory bank would. We also vary the number of components $K$ and the explained variance ratio $\tau$. Finally, we extend the model to a mixture of probabilistic PCA (MPPCA) and test whether its density terms and a noise variance estimated per component improve over the residual score (see \cref{sec:probabilistic}i).

Our core contributions are i) LocalSubspaceAD, a training-free local PCA model for few-shot AD that improves over a single subspace on MVTec AD, VisA and Real-IAD in 1-, 2-, and 4-shot scenarios using both DINOv2 and DINOv3 as backbone and ii) an analysis showing that the improvements stem from locally oriented subspaces fit to semantically coherent regions of feature space.

\section{Related Work}
\label{sec:related_works}

\paragraph{Industrial anomaly detection.}
Visual inspection is commonly automated with unsupervised AD methods trained on nominal images only \cite{liuDeepIndustrialImage2024}, evaluated on benchmarks with pixel-level defect annotations such as MVTec AD \cite{bergmannMVTecAnomalyDetection2021}, VisA \cite{zouSPottheDifferenceSelfsupervisedPretraining2022} and Real-IAD \cite{wangRealIADRealWorldMultiView2024}. With strong pre-trained feature extractors, feature-embedding methods have become favored over reconstruction-based ones \cite{liSurveyDeepLearning2025}. Teacher-student approaches \cite{batznerEfficientADAccurateVisual2024,bergmannUninformedStudentsStudentTeacher2020, zhangDeSTSegSegmentationGuided2023} distill a pre-trained extractor into a student on nominal data, one-class classifiers \cite{ruffDeepOneClassClassification2018a, shangDeepSupportVector2025, yiPatchSVDDPatchLevel2021} learn a boundary around nominal embeddings, and distribution-based methods model nominal features with Gaussians \cite{defardPaDiMPatchDistribution2021,rippelGaussianAnomalyDetection2021, wanIndustrialImageAnomaly2022, yaoHierarchicalGaussianMixture2024} or normalizing flows \cite{gudovskiyCFLOWADRealTimeUnsupervised2022, rudolphSameSameDifferNet2021}. Most of these are trained or estimated on more data than a few-shot setting provides. Memory-bank methods \cite{cohenSubImageAnomalyDetection2021,dammAnomalyDINOBoostingPatchbased2025,guUniVADTrainingfreeUnified2025,rothTotalRecallIndustrial2022,tangDuoADLeveragingCLS2026} are training-free and score a patch by its distance to the nearest stored nominal feature, but with few shots the bank is sparsely populated, so nominal patches with high variability can score highly in such scenarios, resulting in false positives.

\paragraph{Few-shot AD.}
Few-shot anomaly detection limits the number of nominal training images, commonly to 1, 2 or 4. Cross-category training can be used to enable few-shot performance \cite{huangRegistrationBasedFewShot2022, huangFewShotAnomalyDetection2025}. Vision-language-model-based approaches using text prompts or in-context residuals show strong few-shot performance but require prompts or in-context residuals in addition to the few-shot samples \cite{jeongWinCLIPZeroFewShot2023, liPromptADLearningPrompts2024, zhuGeneralistAnomalyDetection2024}.
AnomalyDINO \cite{dammAnomalyDINOBoostingPatchbased2025} showed that kNN on DINOv2 \cite{oquabDINOv2LearningRobust2023} patch features alone is a strong training-free few-shot baseline, which was further improved by FastRef \cite{liFastRefFastPrototype2026} through prototype refinement and DuoAD \cite{tangDuoADLeveragingCLS2026} through class-token-guided re-weighting. UniVAD \cite{guUniVADTrainingfreeUnified2025} partitions images into components using segmentation models and restricts nearest-neighbor matching to reference patches from the same component, complemented by a graph-based model for logical anomalies. Its gains indicate that scoring each patch only against nominal patches from the same component is beneficial, which motivates partitioning the feature space in our approach.

\paragraph{Subspace models.}
Closest to this work, statistical approaches model nominal feature distributions parametrically \cite{defardPaDiMPatchDistribution2021,lenderingSubspaceADTrainingFreeFewShot2026,rippelGaussianAnomalyDetection2021}. Ndiour \etal \cite{ndiourSubspaceModelingFast2022} used the PCA reconstruction error of deep features for out-of-distribution and anomaly detection, and SubspaceAD \cite{lenderingSubspaceADTrainingFreeFewShot2026} brought this idea to few-shot industrial AD with foundation-model features, using a single global subspace. A single subspace has limited capacity to model heterogeneous nominal data. At the other end of the parameter spectrum, Gaussian Mixture Models fit a full covariance per component, which requires estimating many parameters (e.g. PNFDM \cite{fuPatchWiseNeighborhoodFeature2024},  GLAD \cite{artolaGLADGlobaltoLocalAnomaly2023} and DAGMM \cite{zongDeepAutoencodingGaussian2018}). 
Between these two extremes, mixtures of local linear models approximate the nominal distribution with several low-dimensional subspaces. Such models have been studied in various settings, but have not yet been applied to visual anomaly detection with deep features. Kambhatla and Leen \cite{kambhatlaDimensionReductionLocal1997} partition the data into regions using vector quantization and fit a local PCA to each region (VQPCA), minimizing reconstruction error with hard assignments and without a density model. Ghahramani and Hinton \cite{ghahramaniEMAlgorithmMixtures1996} use a mixture of factor analyzers (MFA) to formulate local subspaces with diagonal noise fit by expectation maximization (EM). Tipping and Bishop~\cite{tippingMixturesProbabilisticPrincipal1999} restrict the noise to be isotropic, resulting in mixtures of PPCA (MPPCA) with closed-form per-component estimates. 

\section{Method}
\label{sec:method}

\begin{figure*}[t]
\centering
\includegraphics[width=\linewidth]{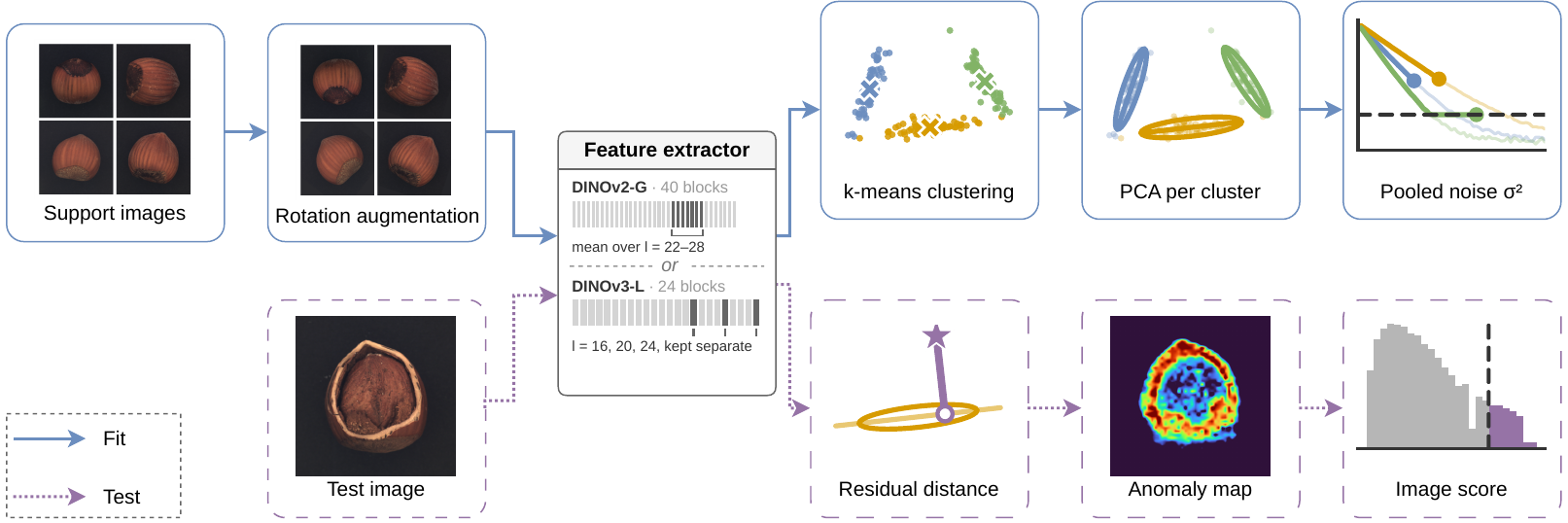}
\caption{Overview of LocalSubspaceAD. Fit (blue): the support images are augmented by rotation, their patch features are clustered with k-means, a PCA is fitted to each cluster and the residual variance is pooled into $\sigma^2$. Test (purple): each patch of the test image is scored by its residual distance to the best matching local subspace, giving the anomaly map, whose top $1\,\%$ form the image score. The feature extractor is either DINOv2-G with layers averaged or DINOv3-L with layers kept separate.}
\label{fig:architecture}
\end{figure*}

\subsection{Problem Setup and Overview}
The goal is to fit a model on a limited set of nominal training images that assigns each patch of a test image an anomaly score $s$, such that anomalous patches $x_a$ score higher than nominal patches $x_n$ such that $s(x_a) > s(x_n)$. The underlying assumption of subspace-based approaches is that the embeddings of nominal patches vary along only a few principal directions, so a feature vector that leaves the subspace spanned by these directions is likely anomalous. We fit one lower dimensional subspace per group of similar patches, so that each semantic region is represented by a specialized subspace. 

\Cref{fig:architecture} shows the resulting pipeline, with the fit path in blue and the test path in purple. Patch embeddings are computed with a foundation model (DINOv2 or DINOv3) for all available training images and partitioned into clusters of similar patches. Each cluster is modeled as a local subspace using PCA. A test patch is scored by its re-projection residual from the subspace that best reconstructs it, and the patch scores are aggregated into an image-level score.

\subsection{Patch Embeddings} 
We follow the feature extraction process used in SubspaceAD \cite{lenderingSubspaceADTrainingFreeFewShot2026} in the DINOv2 based variant (feature extractor in \cref{fig:architecture}). To compute anomaly scores, we first compute a dense feature map $F \in \mathbb{R}^{h \times w \times D}$ with height $h$ and width $w$ where each patch embedding $x_p$ at each spatial position $p$ has dimension $D$. Let $f_l(p)$ denote the intermediate feature vector output by the feature extractor model (DINOv2-G) at layer $l$. $x_p$ is computed as the per-channel average of $f_l(p)$ of a set of layers $\mathcal{L}_\text{avg}=\{22,\dots,28\}$, with $x_p=\frac{1}{|\mathcal{L}_\text{avg}|}\sum_{l \in \mathcal{L}_\text{avg}}f_l(p)$. We follow SubspaceAD's augmentation protocol and rotate images 30 times, so that variation caused by rotation is part of the nominal data. Rotation-sensitive categories, where the orientation itself can be the defect, are not augmented (i.e.: transistor in MVTec AD in the experiments).

For DINOv3 \cite{simeoniDINOv32025b} we keep the layers separate instead of averaging them, because the individual layers are more self-contained (see \cite{simeoniDINOv32025b}). The DINOv3 variant of LocalSubspaceAD builds local subspaces for each layer $l \in \mathcal{L} = \{16, 20, 24\}$ of DINOv3-L independently, and only the scores are aggregated into a single patch score (\cref{eq:scoring_layerwise}). In this variant, the embedding $x_p^{(l)} = f_l(p)$ and all quantities fitted to it below carry an additional layer index $l$, which we omit unless it is needed.

\subsection{Limitations of a Single Subspace}
Subspace-based AD models (such as SubspaceAD) score a patch by its residual, the part of the embedding that is not explained by the subspace. Semantic regions can differ in two ways: i) the position of their embeddings (their means), and ii) the directions along which they vary (the orientation of their covariances). Variations inside the subspace are invisible to the reprojection, so the model's sensitivity is limited by its subspace capacity. A single global subspace must capture the nominal variation of all semantic regions of an object, even though this variation need not be oriented the same way in every region. Suppose a direction $v$ carries enough of region $j$’s variance to be included in the global subspace, but is orthogonal to the local subspace of another region $k$.  A patch from region $k$ that deviates along direction $v$ is not penalized by the global model  because $v$ lies inside the global subspace and is therefore reconstructed. A local subspace for semantic region $k$ penalizes this deviation since $v$ is not part of its subspace.

We therefore model the patch embeddings with a mixture of $K$ local subspaces with index $k$, so that each patch is measured only against the variation of its own semantic region. Assigning each semantic region a local subspace allows us to: i) model the natural variation in appearance within a region more tightly than a single global subspace, ii) treat a direction of variation as nominal in one region without accepting it in others, and iii) choose the number of retained principal directions per subspace according to the variance explained within that subspace. 

The law of total covariance separates two types of variation. For any partition of the patches into $K$ groups, with $\pi_k$ the fraction of patches in group $k$, $\mu_k$ its empirical mean and $\Sigma_k$ its covariance, the total covariance splits into a within- and a between-group term, $\Sigma_W$ and $\Sigma_B$, respectively,
\begin{equation} \label{eq:covariance_model}
\begin{split}
    \Sigma   &= \Sigma_W + \Sigma_B, \qquad \Sigma_W = \sum_{k=1}^{K} \pi_k \Sigma_k, \\
    \Sigma_B &= \sum_{k=1}^{K} \pi_k (\mu_k - \mu)(\mu_k - \mu)^\mathsf{T},
\end{split}
\end{equation}
where $\mu$ is the empirical mean of all embeddings $x_p$. $\Sigma_B$ has rank at most $K-1$, so the cluster means can add at most $K-1$ dimensions of variation to the global subspace. The directions add far more. When each local subspace is constructed to explain $99\,\%$ of the variance of its corresponding semantic region (the threshold $\tau$ in \cref{eq:tau}) fewer directions are needed compared to a single global subspace. Our experiments show a single global subspace requires $r \approx 940$ directions, against a median of $r_k \approx 312$ per local subspace at $K=128$ (\cref{sec:supp_dims}). 

Every direction retained in a global subspace may cost sensitivity for some semantic region, since it is a direction along which deviations of that region go unpenalized. The local offsets matter little by comparison to the principal directions: centering every component on the global mean $\mu$ instead of $\mu_k$ matches the full model at 1-shot and is slightly better at 4-shot (at most $+0.18$ points) (\emph{shared mean} in \cref{tab:ablation_paired}).

\subsection{Mixture of Local Subspaces}

Each patch should be measured only against the variation of the corresponding semantic region. Patches that belong to the same semantic region of an object or that have the same texture have similar embeddings, so we group the embeddings of the training images with k-means clustering, which needs no labels or masks, and fit one PCA per cluster (\emph{k-means clustering} and \emph{PCA per cluster} in \cref{fig:architecture}). Let $c_p \in \{1, \dots, K\}$ denote the cluster assigned to $x_p$, and let $X_k = \{x_p \mid c_p = k \,\}$ denote the patch embeddings assigned to cluster $k$, with $N = \sum_k |X_k|$ support patches in total, so that $\pi_k = | X_k| / N$. We fit a PCA to each cluster's embeddings $X_k$, computing the empirical mean $\mu_k$ and the eigenvalues $\lambda_{k,1} \geq \dots \geq \lambda_{k,D}$ of $\Sigma_k$. The eigenvectors of the largest $r_k$ eigenvalues form the matrix $B_k \in \mathbb{R}^{D \times r_k}$ whose orthonormal columns span the local subspace of cluster $k$. The choice of $r_k$ is given in \cref{eq:tau}. This gives the following model of the data 

\begin{equation}
    \label{eq:local_pca}
    x_p = \mu_k + B_k z_p + e_p, \qquad k = c_p
\end{equation}
with $B_k^\mathsf{T} B_k = I_{r_k}$, $z_p = B_k^\mathsf{T} (x_p - \mu_k)$ and $e_p = (I_D - B_k B_k^\mathsf{T})(x_p - \mu_k)$ orthogonal to the subspace. Here $z_p$ holds the values of the feature vector along the directions in which its cluster normally varies, and $e_p$ is the remainder that the cluster cannot explain; only $e_p$ counts as evidence for an anomaly. For a given cluster assignment and dimension $r_k$, $B_k$ minimizes the reconstruction error $\sum_{x_p \in X_k} \| e_p\| ^2$ of its cluster (Eckart-Young \cite{eckartApproximationOneMatrix1936}), so each local subspace is the best linear summary of its cluster. The clustering itself is not optimized for reconstruction: patches are assigned by Euclidean distance to the cluster centers, not by their residual to each subspace. This corresponds to the Euclidean variant of VQPCA~\cite{kambhatlaDimensionReductionLocal1997}.
The projection into the subspace of component $k$ is 
\begin{equation} \label{eq:projection}
    x_{p,k}^\text{proj}=\mu_k + B_kB_k^\mathsf{T}(x_p-\mu_k).
\end{equation}

In the studied scenarios, the clusters differ in complexity: a uniform background varies along few dimensions, a textured surface along many. A fixed number of dimensions would be too permissive for the former or too strict for the latter. $r_k$ is therefore chosen per component as the smallest number of dimensions that explains a fraction $\tau$ of the component's variance (as in SubspaceAD):
\begin{equation}
\label{eq:tau}
r_k = \min \Big\{ r \mid \frac{\sum_{j=1}^{r} \lambda_{k,j}}{\sum_{j=1}^{D} \lambda_{k,j}} \geq \tau \Big\}.
\end{equation}

SubspaceAD is the special case of our model with a single cluster ($K = 1$) on the DINOv2 variant.

All layers share the same cluster assignments (computed on layer $20$) in the DINOv3 variant. Empirical evaluation shows the best results with a shared layer selection (see \cref{sec:layer-selection}). $K$ and $\tau$ are hyperparameters (experiments use $K=128$ and $\tau = 0.99$). $K$ controls how specific each local subspace is: more components give smaller and stricter subspaces, but leave fewer patches to estimate each of them. The choice of $K$ is examined in \cref{ssec:k-sweep}. Since a PCA fit to very few patches is unreliable, $K$ is capped at $\lfloor N / 20 \rfloor$ so that components hold at least 20 patches on average.

The variance not modeled by the subspaces determines how large the residual of a nominal patch is. We summarize it by a global residual (noise) variance $\sigma^2$ shared by all components (\emph{pooled noise} in \cref{fig:architecture}). It is only used in the DINOv3 variant to normalize the layers for scoring (\cref{eq:scoring_layerwise}). Rather than estimating it for each component separately, which is unreliable for components with a small number of patches, we pool the residual variance and the number of discarded directions across all components, weighted by component size:
\begin{equation}\label{eq:sigma_pooled}
    \sigma^2 = \frac{\sum_k |X_k| \sum_{j > r_k}^{D} \lambda_{k,j}}
                    {\sum_k |X_k| \max(R_k - r_k, 1)},
\end{equation}
where $R_k = \min(|X_k| - 1, D)$ is the maximal rank of $\Sigma_k$. Using $R_k$ instead of $D$ avoids counting the $D-R_k$ directions without sample variance as noise, which would otherwise shrink $\sigma^2$ when $|X_k| < D$. In the DINOv3 variant, $\sigma_l^2$ is computed per layer $l$ in the same way.

\subsection{Scoring} 
\label{sec:scoring}
For a test patch it is unknown which semantic region it belongs to, and an anomalous patch may belong to none. We therefore do not assign patches to a cluster but compare it against every component: the anomaly score of a patch embedding $x_p$ should be high only if no local subspace can reconstruct it. Consequently, we use the residual distance to the best matching component (\emph{residual distance} in \cref{fig:architecture}) to compute

\begin{equation}
\label{eq:scoring}
s(x_p) = \min_k d_k(x_p), \qquad d_k(x_p) = \| x_p - x_{p,k}^{\text{proj}}\|^2.
\end{equation}

In the layer-wise DINOv3 variant each component has one subspace per layer, since already at $K=1$ a one subspace per layer performs better than a single subspace on the layer-averaged features (DINOv3 rows in \cref{tab:ablation_paired}). The normalized residuals of a component are summed over all layers before the minimum across components is taken:
\begin{equation}
\label{eq:scoring_layerwise}
s(x_p) = \min_k \sum_{l \in \mathcal{L}} d_{l,k}\big(x_p^{(l)}\big) / \sigma_l^2
\end{equation}

with $d_{l,k}$ being the residual distance of the layer-$l$ embedding $x_p^{(l)}$ to the subspace fitted to component $k$ at layer $l$ and $\sigma_l^2$ the shared noise variance \cref{eq:sigma_pooled} of layer $l$. Dividing by $\sigma_l^2$ expresses each residual in units of the nominal residual variance of its layer. Without this normalization the scales of the layers differ and the sum would be dominated by one layer (\cref{tab:ppca_decomposition}).

The rest of the scoring is unchanged from SubspaceAD (\emph{anomaly map} and \emph{image score} in \cref{fig:architecture}). The patch scores $s(x_p)$ are bilinearly upsampled to a pixel-level map and blurred with a $3 \times 3$ Gaussian blur with $\sigma_{\text{blur}} = 4$ (close to a box filter). The resulting pixel-level scores are aggregated into an image-level score by computing the mean of their top $\rho=1$ \%, since defects usually cover only a small part of the image and would be averaged out by a mean over all pixels. In the DINOv3 variant, image-level scores are computed without the two outermost rows and columns of patch embeddings, because artifacts of the backbone cause high anomaly scores in some border patches.

\subsection{Computational Cost}

The additional computational cost compared to SubspaceAD's scoring head grows linearly with $K$ at inference time and stays small compared to feature extraction; fitting adds one k-means run and one PCA for \textbf{$K$} components. During inference, the cost is dominated by the projections $B_k^\mathsf{T}(x_p - \mu_k)$. The residual can be computed without full re-projection, with 
\begin{equation}\label{eq:score_fast}
d_k(x_p) = \|x_p - \mu_k\|^2 - \|B_k^\mathsf{T}(x_p - \mu_k)\|^2.
\end{equation}
SubspaceAD costs $\mathcal{O}(D r)$ dot products per patch, in contrast LocalSubspaceAD costs $\mathcal{O}(D \sum_{k=1}^K r_k)$ per patch (a factor of roughly $K \tilde{r} / r$, with $\tilde{r}$ the median $r_k$). With $D = 1536$, $K=128$ and $\tilde{r}\approx 312$ (\cref{sec:supp_dims}) that is about 120 MFLOPs per patch, or about 5 to 10~\% on top of the DINOv2 backbone. The residual computation can be parallelized in time critical applications, further reducing the impact.


\section{Experiments}
\label{sec:results}

\begin{figure}[t]
\centering
\includegraphics[width=\columnwidth]{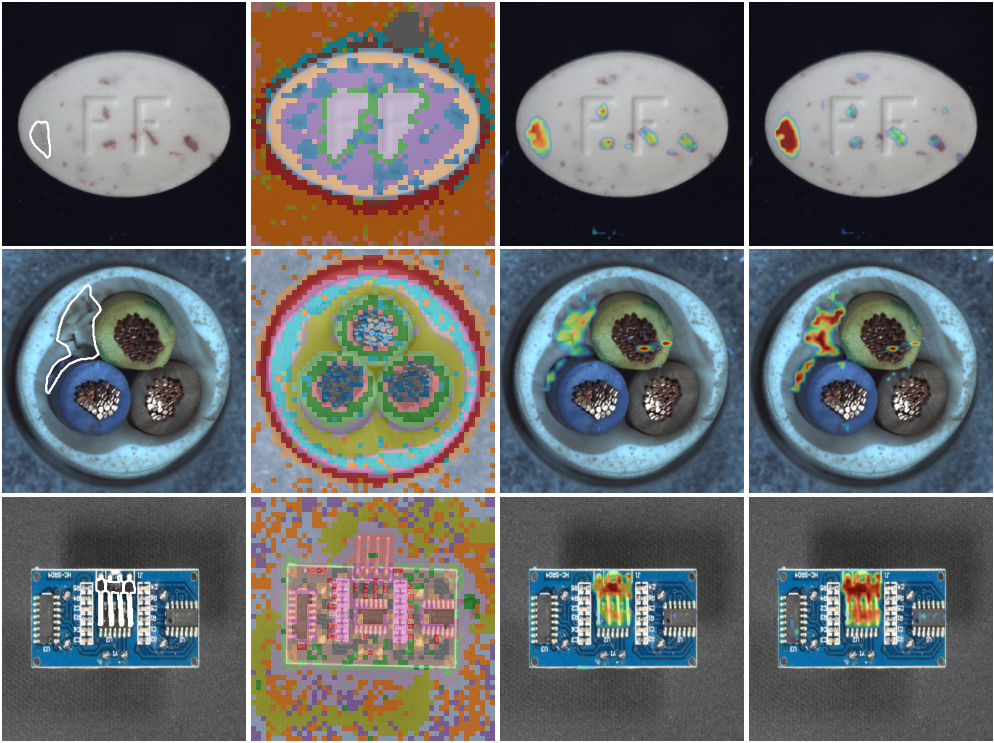}
\caption{Qualitative results on MVTec AD (pill, cable) and VisA (pcb2). Left to right: test image with ground-truth outline, assigned component per patch of a nominal image ($K$$=$$16$), anomaly maps of SubspaceAD and LocalSubspaceAD ($K$$=$$128$).}
\label{fig:qualitative}
\end{figure}

\subsection{Setup}
We evaluate on MVTec AD, VisA and Real-IAD (single view, C1) in the 1-, 2- and 4-shot settings and report image-level AUROC, pixel-level AUPR, and AUPRO. Each result is the mean over categories, averaged over 10 seeds; each seed draws a different support set, and all methods use the same support sets per seed. $\pm$ denotes the standard deviation over seeds. Notice that our results deviate slightly from those reported in the original papers, this is attributed to the differences in random seeding.

The sensitivity analyses and ablations (\cref{ssec:k-sweep,ssec:ablation}) use three of these seeds on VisA with DINOv2 unless stated otherwise. SubspaceAD, AnomalyDINO, DuoAD and UniVAD are re-run from the respective official implementations using the same support samples on all methods. Unless stated otherwise LocalSubspaceAD uses $K = 128$ and $\tau = 0.99$ throughout, with $K$ lowered to the cap $\lfloor N/20 \rfloor$ of \cref{sec:method} where the support set is too small. The hyperparameters were selected on MVTec AD categories and used unchanged on VisA and Real-IAD. The sweeps on VisA in \cref{ssec:k-sweep} are a post-hoc sensitivity analysis and were not used for selection.

\begin{table*}[t]
\centering
\caption{Few-shot anomaly detection, mean $\pm$ std over 10 seeds. Best per dataset and shot count in \textbf{bold}, second \underline{underlined}. LocalSubspaceAD is above SubspaceAD on the same backbone at every seed, except where marked $^{\circ}$ (\cref{tab:fewshot_summary,tab:fewshot_paired}). DINOv2: ViT-g/14 with registers; DINOv3: ViT-L/16, with SubspaceAD on the mean of layers 16, 20 and 24. UniVAD~\cite{guUniVADTrainingfreeUnified2025} uses its own backbones and was not run on Real-IAD due to computational costs. AnomalyDino uses DINOv2-S. $^{\ddagger}$As reported by the original authors; --: not reported. All metrics and UniVAD's backbones are in \cref{tab:fewshot_full}.}
\label{tab:fewshot}
\setlength{\tabcolsep}{2.5pt}
\footnotesize
\begin{tabular}{cll@{\hspace{5pt}}ccc@{\hspace{5pt}}ccc@{\hspace{5pt}}ccc}
\toprule
& & & \multicolumn{3}{c}{MVTec AD} & \multicolumn{3}{c}{VisA} & \multicolumn{3}{c}{Real-IAD} \\
\cmidrule(lr){4-6} \cmidrule(lr){7-9} \cmidrule(lr){10-12}
Shots & Backbone & Method & I-AUROC & P-AUPR & AUPRO & I-AUROC & P-AUPR & AUPRO & I-AUROC & P-AUPR & AUPRO \\
\midrule
\multirow{9}{*}{1} & \multirow{2}{*}{Reported} & AnomalyDino+~\cite{liFastRefFastPrototype2026}$^{\ddagger}$ & \underline{97.00} & -- & -- & 84.60 & -- & -- & -- & -- & -- \\
 &  & WinCLIP~\cite{jeongWinCLIPZeroFewShot2023}$^{\ddagger}$ & 93.10{\tiny$\pm$2.00} & -- & -- & 83.80{\tiny$\pm$4.00} & -- & -- & -- & -- & -- \\
\addlinespace[3pt]
 & \multirow{2}{*}{Other} & AnomalyDINO & 95.77{\tiny$\pm$0.93} & 54.92{\tiny$\pm$1.16} & 91.38{\tiny$\pm$0.36} & 88.44{\tiny$\pm$1.44} & 35.73{\tiny$\pm$0.82} & 91.14{\tiny$\pm$0.33} & 84.54{\tiny$\pm$0.62} & 32.57{\tiny$\pm$1.03} & 92.44{\tiny$\pm$0.39} \\
 &  & UniVAD & 96.06{\tiny$\pm$1.31} & 54.23{\tiny$\pm$0.75} & 92.64{\tiny$\pm$0.19} & 93.16{\tiny$\pm$0.84} & 36.73{\tiny$\pm$0.58} & 92.64{\tiny$\pm$0.43} & -- & -- & -- \\
\addlinespace[3pt]
 & \multirow{2}{*}{DINOv2} & SubspaceAD & 96.04{\tiny$\pm$1.38} & 58.26{\tiny$\pm$0.95} & 94.38{\tiny$\pm$0.16} & 92.55{\tiny$\pm$0.79} & 36.96{\tiny$\pm$0.75} & \underline{95.40}{\tiny$\pm$0.11} & \underline{88.60}{\tiny$\pm$0.50} & 44.02{\tiny$\pm$0.90} & \underline{97.40}{\tiny$\pm$0.21} \\
 &  & \textbf{LocalSubspaceAD} & \textbf{97.33}{\tiny$\pm$0.84} & 58.72{\tiny$\pm$1.09}\rlap{$^{\circ}$} & \underline{94.77}{\tiny$\pm$0.20} & \textbf{94.00}{\tiny$\pm$0.74} & 41.82{\tiny$\pm$0.81} & \textbf{95.71}{\tiny$\pm$0.12} & \textbf{90.88}{\tiny$\pm$0.36} & 48.72{\tiny$\pm$0.85} & \textbf{97.77}{\tiny$\pm$0.16} \\
\addlinespace[3pt]
 & \multirow{3}{*}{DINOv3} & DuoAD & 96.59{\tiny$\pm$0.86} & 63.63{\tiny$\pm$0.93} & 93.83{\tiny$\pm$0.32} & 92.28{\tiny$\pm$0.64} & 45.21{\tiny$\pm$0.39} & 92.80{\tiny$\pm$0.26} & 88.18{\tiny$\pm$0.67} & 50.96{\tiny$\pm$0.85} & 96.25{\tiny$\pm$0.19} \\
 &  & SubspaceAD & 94.97{\tiny$\pm$0.83} & \underline{68.45}{\tiny$\pm$0.89} & 93.37{\tiny$\pm$0.49} & 90.36{\tiny$\pm$0.96} & \underline{50.17}{\tiny$\pm$0.85} & 93.42{\tiny$\pm$0.23} & 83.62{\tiny$\pm$0.98} & \underline{51.14}{\tiny$\pm$1.08} & 94.15{\tiny$\pm$0.34} \\
 &  & \textbf{LocalSubspaceAD} & 96.97{\tiny$\pm$0.99} & \textbf{70.64}{\tiny$\pm$0.90} & \textbf{95.24}{\tiny$\pm$0.29} & \underline{93.57}{\tiny$\pm$0.58} & \textbf{55.09}{\tiny$\pm$0.59} & 94.83{\tiny$\pm$0.18} & 88.20{\tiny$\pm$0.52} & \textbf{56.70}{\tiny$\pm$0.75} & 96.26{\tiny$\pm$0.19} \\
\cmidrule(lr){1-12}
\multirow{9}{*}{2} & \multirow{2}{*}{Reported} & AnomalyDino+~\cite{liFastRefFastPrototype2026}$^{\ddagger}$ & 97.20 & -- & -- & 84.80 & -- & -- & -- & -- & -- \\
 &  & WinCLIP~\cite{jeongWinCLIPZeroFewShot2023}$^{\ddagger}$ & 94.40{\tiny$\pm$1.30} & -- & -- & 84.60{\tiny$\pm$2.40} & -- & -- & -- & -- & -- \\
\addlinespace[3pt]
 & \multirow{2}{*}{Other} & AnomalyDINO & 96.49{\tiny$\pm$0.64} & 55.90{\tiny$\pm$0.84} & 91.97{\tiny$\pm$0.27} & 90.71{\tiny$\pm$1.04} & 38.21{\tiny$\pm$0.67} & 92.05{\tiny$\pm$0.23} & 86.79{\tiny$\pm$0.34} & 33.73{\tiny$\pm$0.67} & 93.39{\tiny$\pm$0.25} \\
 &  & UniVAD & 96.99{\tiny$\pm$1.08} & 55.50{\tiny$\pm$0.50} & 93.12{\tiny$\pm$0.15} & 94.77{\tiny$\pm$0.53} & 39.24{\tiny$\pm$0.45} & 93.40{\tiny$\pm$0.25} & -- & -- & -- \\
\addlinespace[3pt]
 & \multirow{2}{*}{DINOv2} & SubspaceAD & 96.76{\tiny$\pm$0.91} & 59.96{\tiny$\pm$0.71} & 94.83{\tiny$\pm$0.11} & 93.99{\tiny$\pm$0.59} & 39.39{\tiny$\pm$0.83} & \underline{95.77}{\tiny$\pm$0.08} & \underline{90.30}{\tiny$\pm$0.42} & 45.67{\tiny$\pm$0.59} & \underline{97.69}{\tiny$\pm$0.09} \\
 &  & \textbf{LocalSubspaceAD} & \textbf{98.00}{\tiny$\pm$0.44} & 60.43{\tiny$\pm$0.53}\rlap{$^{\circ}$} & \underline{95.30}{\tiny$\pm$0.27} & \textbf{95.20}{\tiny$\pm$0.53} & 43.76{\tiny$\pm$0.79} & \textbf{96.20}{\tiny$\pm$0.12} & \textbf{92.09}{\tiny$\pm$0.34} & 49.95{\tiny$\pm$0.59} & \textbf{98.01}{\tiny$\pm$0.07} \\
\addlinespace[3pt]
 & \multirow{3}{*}{DINOv3} & DuoAD & 97.20{\tiny$\pm$0.70} & 64.96{\tiny$\pm$0.84} & 94.45{\tiny$\pm$0.31} & 93.10{\tiny$\pm$0.61} & 47.50{\tiny$\pm$0.50} & 93.80{\tiny$\pm$0.17} & 89.85{\tiny$\pm$0.51} & 52.41{\tiny$\pm$0.52} & 96.73{\tiny$\pm$0.07} \\
 &  & SubspaceAD & 96.07{\tiny$\pm$0.48} & \underline{69.23}{\tiny$\pm$0.53} & 94.01{\tiny$\pm$0.32} & 91.84{\tiny$\pm$0.58} & \underline{51.67}{\tiny$\pm$0.68} & 94.00{\tiny$\pm$0.11} & 86.04{\tiny$\pm$0.69} & \underline{52.88}{\tiny$\pm$0.80} & 95.13{\tiny$\pm$0.15} \\
 &  & \textbf{LocalSubspaceAD} & \underline{97.48}{\tiny$\pm$0.83} & \textbf{71.06}{\tiny$\pm$0.83} & \textbf{95.72}{\tiny$\pm$0.28} & \underline{94.80}{\tiny$\pm$0.54} & \textbf{55.53}{\tiny$\pm$0.42} & 95.46{\tiny$\pm$0.09} & 89.91{\tiny$\pm$0.41} & \textbf{57.33}{\tiny$\pm$0.50} & 96.77{\tiny$\pm$0.11} \\
\cmidrule(lr){1-12}
\multirow{9}{*}{4} & \multirow{2}{*}{Reported} & AnomalyDino+~\cite{liFastRefFastPrototype2026}$^{\ddagger}$ & 97.40 & -- & -- & 87.10 & -- & -- & -- & -- & -- \\
 &  & WinCLIP~\cite{jeongWinCLIPZeroFewShot2023}$^{\ddagger}$ & 95.20{\tiny$\pm$1.30} & -- & -- & 87.30{\tiny$\pm$1.80} & -- & -- & -- & -- & -- \\
\addlinespace[3pt]
 & \multirow{2}{*}{Other} & AnomalyDINO & 97.39{\tiny$\pm$0.56} & 56.80{\tiny$\pm$0.74} & 92.40{\tiny$\pm$0.27} & 92.22{\tiny$\pm$0.70} & 39.72{\tiny$\pm$0.37} & 92.77{\tiny$\pm$0.33} & 88.89{\tiny$\pm$0.49} & 35.47{\tiny$\pm$0.78} & 94.22{\tiny$\pm$0.21} \\
 &  & UniVAD & 97.99{\tiny$\pm$0.89} & 56.55{\tiny$\pm$0.63} & 93.43{\tiny$\pm$0.12} & 95.75{\tiny$\pm$0.42} & 40.66{\tiny$\pm$0.49} & 93.78{\tiny$\pm$0.10} & -- & -- & -- \\
\addlinespace[3pt]
 & \multirow{2}{*}{DINOv2} & SubspaceAD & 97.66{\tiny$\pm$0.70} & 61.38{\tiny$\pm$0.82} & 95.16{\tiny$\pm$0.09} & 94.79{\tiny$\pm$0.43} & 40.66{\tiny$\pm$0.47} & \underline{96.00}{\tiny$\pm$0.12} & \underline{91.65}{\tiny$\pm$0.40} & 47.44{\tiny$\pm$0.30} & \underline{97.96}{\tiny$\pm$0.06} \\
 &  & \textbf{LocalSubspaceAD} & \textbf{98.58}{\tiny$\pm$0.44} & 61.41{\tiny$\pm$0.91}\rlap{$^{\circ}$} & \underline{95.70}{\tiny$\pm$0.16} & \textbf{96.21}{\tiny$\pm$0.39} & 45.00{\tiny$\pm$0.96} & \textbf{96.55}{\tiny$\pm$0.11} & \textbf{93.03}{\tiny$\pm$0.38} & 50.65{\tiny$\pm$0.19} & \textbf{98.23}{\tiny$\pm$0.12} \\
\addlinespace[3pt]
 & \multirow{3}{*}{DINOv3} & DuoAD & 98.01{\tiny$\pm$0.65} & 66.65{\tiny$\pm$0.88} & 95.07{\tiny$\pm$0.19} & 93.82{\tiny$\pm$0.35} & 49.03{\tiny$\pm$0.46} & 94.55{\tiny$\pm$0.11} & 91.17{\tiny$\pm$0.35} & 53.86{\tiny$\pm$0.41} & 97.08{\tiny$\pm$0.05} \\
 &  & SubspaceAD & 97.30{\tiny$\pm$0.65} & \underline{70.54}{\tiny$\pm$0.99} & 94.55{\tiny$\pm$0.25} & 93.11{\tiny$\pm$0.43} & \underline{52.47}{\tiny$\pm$0.62} & 94.31{\tiny$\pm$0.09} & 87.69{\tiny$\pm$0.46} & \underline{54.25}{\tiny$\pm$0.56} & 95.72{\tiny$\pm$0.09} \\
 &  & \textbf{LocalSubspaceAD} & \underline{98.31}{\tiny$\pm$0.74} & \textbf{71.99}{\tiny$\pm$0.91} & \textbf{96.26}{\tiny$\pm$0.15} & \underline{95.91}{\tiny$\pm$0.51} & \textbf{54.94}{\tiny$\pm$0.39} & 95.69{\tiny$\pm$0.12} & 90.99{\tiny$\pm$0.25} & \textbf{57.36}{\tiny$\pm$0.35} & 97.12{\tiny$\pm$0.08} \\
\bottomrule
\end{tabular}
\end{table*}

\subsection{Few-shot anomaly detection}
\Cref{tab:fewshot} compares LocalSubspaceAD with SubspaceAD and other recent few-shot methods. In the 1-, 2- and 4-shot setting, LocalSubspaceAD provides the best result of all methods considered on every metric and dataset, with the backbone determining on which metric. With DINOv2 it reaches the highest image-level AUROC: $97.33$, $94.00$ and $90.88$ at 1-shot and $98.58$, $96.21$ and $93.03$ at 4-shot on MVTec AD, VisA and Real-IAD. With DINOv3 it reaches the highest pixel-level AUPR: $70.64$, $55.09$ and $56.70$ at 1-shot and $71.99$, $54.94$ and $57.36$ at 4-shot. The highest AUPRO is obtained with DINOv3 on MVTec AD and with DINOv2 on VisA and Real-IAD. The two backbones differ in their strengths, and this difference is a property of the backbone configuration rather than of the model fitted to them: SubspaceAD and LocalSubspaceAD alike reach a higher image-level AUROC with DINOv2 and a higher pixel-level AUPR with DINOv3, on every dataset and at every shot count. In pixel-level AUPR the DINOv2 variants consequently trail all DINOv3-based methods including DuoAD, on every dataset. The local subspaces leave this split unchanged: on average, LocalSubspaceAD improves anomaly detection and localization performance over SubspaceAD in every setting on either backbone. We therefore compare methods on the same backbone. 

The variation in performance caused by the choice of support set is often as large as the performance differences between methods. We therefore also compare methods by paired differences (\cref{tab:fewshot_summary}): per seed, the difference of the two methods' category means on the same support sets, reported as mean $\pm$ std over seeds. A difference with the same sign at every seed and shot count is set in bold. In the few-shot settings, LocalSubspaceAD improves over SubspaceAD at every seed and shot on every metric and dataset, except pixel-level AUPR on MVTec AD with DINOv2 ($+0.32 \pm 0.60$, no clear winner; marked $^{\circ}$ in \cref{tab:fewshot}). On DINOv3 the gain over SubspaceAD is between $+1.42$ and $+4.37$ points on every metric and dataset, and larger than on DINOv2 except for pixel-level AUPR on VisA ($+3.76$ against $+4.52$). This highlights the benefit of modeling the data with multiple subspaces. Against DuoAD on the same DINOv3 ViT-L/16 backbone, LocalSubspaceAD is clearly better on pixel-level AUPR on every dataset ($+4.72$ to $+7.94$), better on AUPRO on MVTec AD and VisA ($+1.29$ and $+1.61$) and on par on Real-IAD ($+0.03 \pm 0.06$). On image-level AUROC it is better on VisA ($+1.69 \pm 0.38$) and on par on MVTec AD and Real-IAD ($+0.32 \pm 0.20$ and $-0.03 \pm 0.19$). The full per-shot differences are in \cref{tab:fewshot_paired}.

\begin{table*}[t]
\centering
\caption{LocalSubspaceAD minus the baselines on the same backbone, in points, averaged over 1/2/4 shots: mean $\pm$ std over 10 seeds of the paired difference. \textbf{Bold}: the same sign at every seed and shot count.}
\label{tab:fewshot_summary}
\setlength{\tabcolsep}{2.5pt}
\footnotesize
\begin{tabular}{ll@{\hspace{5pt}}ccc@{\hspace{5pt}}ccc@{\hspace{5pt}}ccc}
\toprule
& & \multicolumn{3}{c}{MVTec AD} & \multicolumn{3}{c}{VisA} & \multicolumn{3}{c}{Real-IAD} \\
\cmidrule(lr){3-5} \cmidrule(lr){6-8} \cmidrule(lr){9-11}
Backbone & LocalSubspaceAD minus & I-AUROC & P-AUPR & AUPRO & I-AUROC & P-AUPR & AUPRO & I-AUROC & P-AUPR & AUPRO \\
\midrule
DINOv2 & SubspaceAD & \textbf{+1.16}{\tiny$\pm$0.50} & +0.32{\tiny$\pm$0.60} & \textbf{+0.46}{\tiny$\pm$0.10} & \textbf{+1.36}{\tiny$\pm$0.39} & \textbf{+4.52}{\tiny$\pm$0.58} & \textbf{+0.44}{\tiny$\pm$0.08} & \textbf{+1.82}{\tiny$\pm$0.22} & \textbf{+4.06}{\tiny$\pm$0.38} & \textbf{+0.32}{\tiny$\pm$0.06} \\
\addlinespace[3pt]
\multirow{2}{*}{DINOv3} & SubspaceAD & \textbf{+1.48}{\tiny$\pm$0.39} & \textbf{+1.82}{\tiny$\pm$0.34} & \textbf{+1.76}{\tiny$\pm$0.16} & \textbf{+2.99}{\tiny$\pm$0.40} & \textbf{+3.76}{\tiny$\pm$0.54} & \textbf{+1.42}{\tiny$\pm$0.13} & \textbf{+3.91}{\tiny$\pm$0.40} & \textbf{+4.37}{\tiny$\pm$0.38} & \textbf{+1.71}{\tiny$\pm$0.12} \\
 & DuoAD & +0.32{\tiny$\pm$0.20} & \textbf{+6.15}{\tiny$\pm$0.24} & \textbf{+1.29}{\tiny$\pm$0.09} & \textbf{+1.69}{\tiny$\pm$0.38} & \textbf{+7.94}{\tiny$\pm$0.27} & \textbf{+1.61}{\tiny$\pm$0.11} & -0.03{\tiny$\pm$0.19} & \textbf{+4.72}{\tiny$\pm$0.38} & +0.03{\tiny$\pm$0.06} \\
\bottomrule
\end{tabular}
\end{table*}

By design, LocalSubspaceAD should be most beneficial for categories with heterogeneous nominal appearance. To verify this, we group the VisA categories into multi-instance objects, complex PCBs and single-instance objects (\cref{tab:visa_groups}). On multi-instance categories and PCBs LocalSubspaceAD gains $+1.98$ and $+2.78$ I-AUROC and $+8.43$ and $+5.86$ P-AUPR, while on single-instance categories (SubspaceAD is already at $97.60$ I-AUROC) it loses $0.69$ I-AUROC and is on par in localization. These results confirm the design rationale: the gains concentrate on categories with heterogeneous nominal appearance, while categories with homogeneous appearance, where a single subspace already suffices, see no benefit and a small loss in image-level detection.

\textbf{Qualitative results.} \Cref{fig:qualitative} shows anomaly maps for two MVTec AD and one VisA examples. Without any supervision, the component assignment $k^*$ follows the object structure: the embossed letters, body and rim of the pill, the strands, insulation and sheath of the cable and the board, solder pads and mounted chips of the PCB components each fall into separate components, so each semantic region is scored against its own subspace. On the larger, low-contrast defects SubspaceAD responds weakly and only in parts of the annotated region, whereas LocalSubspaceAD responds strongly over most of it. The small color deviation in pill is highlighted less strongly by LocalSubspaceAD than by SubspaceAD.

\begin{table}[t]
\centering
\caption{LocalSubspaceAD minus SubspaceAD on VisA by category group (DINOv2), in points, averaged over 1/2/4 shots: mean $\pm$ std over 10 seeds of the paired difference. Base: SubspaceAD's I-AUROC. Groups and per-category results are in \cref{tab:visa_categories}.}
\label{tab:visa_groups}
\setlength{\tabcolsep}{3.3pt}
\small
\begin{tabular}{lcccc}
\toprule
Group & Base & $\Delta$ I-AUROC & $\Delta$ P-AUPR & $\Delta$ AUPRO \\
\midrule
Multi-instance & 91.72 & +1.98{\scriptsize$\pm$0.76} & +8.43{\scriptsize$\pm$0.55} & +0.45{\scriptsize$\pm$0.15} \\
PCB & 92.01 & +2.78{\scriptsize$\pm$0.89} & +5.86{\scriptsize$\pm$1.28} & +0.81{\scriptsize$\pm$0.26} \\
Single-instance & 97.60 & -0.69{\scriptsize$\pm$0.23} & -0.73{\scriptsize$\pm$0.82} & +0.04{\scriptsize$\pm$0.07} \\
\midrule
All VisA & 93.78 & +1.36{\scriptsize$\pm$0.39} & +4.52{\scriptsize$\pm$0.58} & +0.44{\scriptsize$\pm$0.08} \\
\bottomrule
\end{tabular}
\end{table}

\subsection{\texorpdfstring{Sensitivity to $\boldsymbol{K}$ and $\boldsymbol{\tau}$}{Sensitivity to K and tau}}

\label{ssec:k-sweep}
\begin{figure}[h]
\centering
\includegraphics[width=\columnwidth]{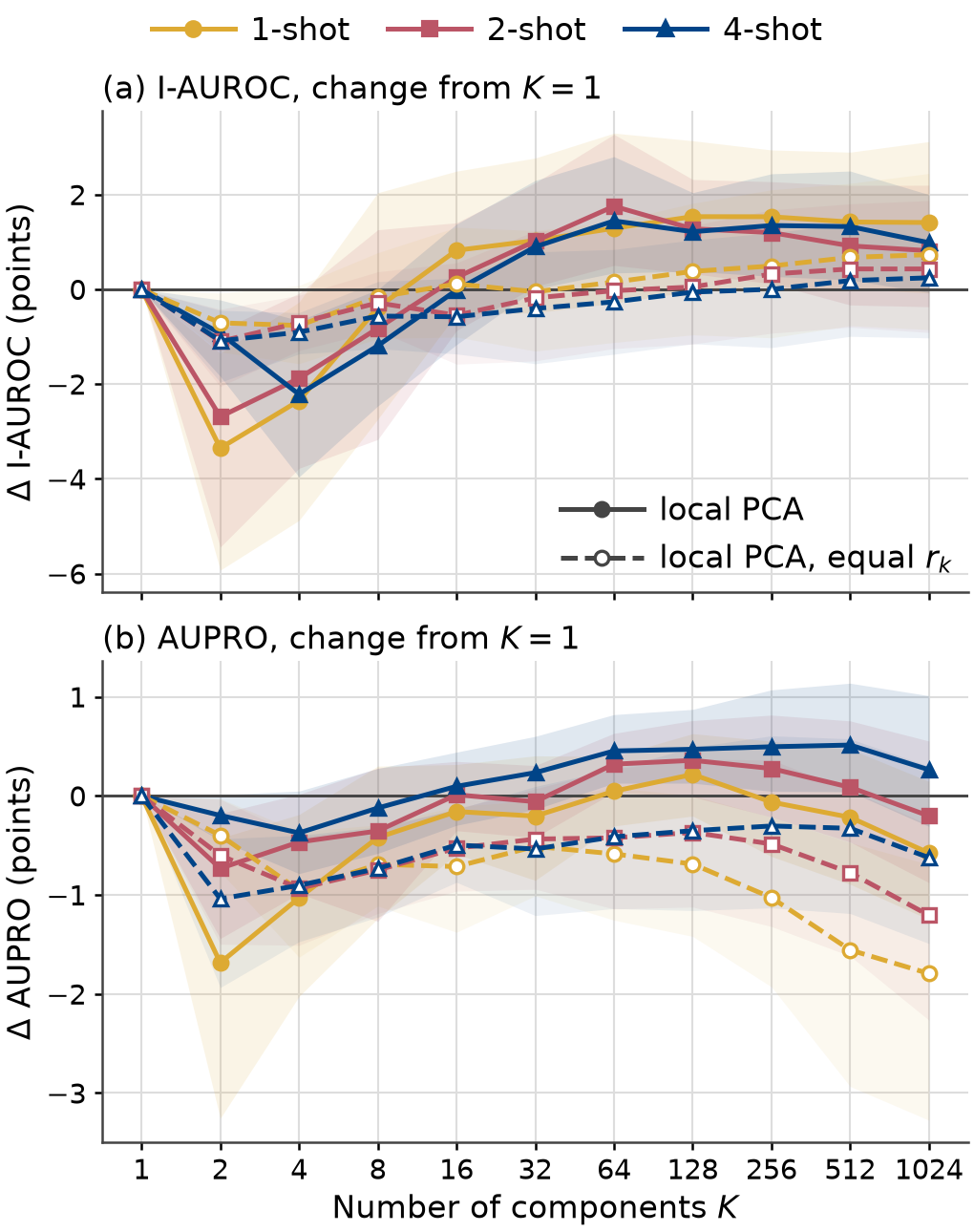}
\caption{Change in image AUROC (a) and AUPRO (b) compared with a single subspace model ($K=1$) at different $K$ components per shot count, VisA, three seeds. Solid: local PCA residual score with per-component $r_k$; dashed: the same number of directions $r_k$ in every component. Band: bootstrapped $90 \%$ CI.}
\label{fig:k_dip}
\end{figure}

\textbf{Number of components $\boldsymbol{K}$.} We vary $K$ from $1$ to $1024$ (\cref{fig:k_dip}, solid blue, and \cref{tab:ppca_ksweep}), capped at $\lfloor N/20 \rfloor$ for $N$ support patches. Averaged over shots, performance first drops below a single subspace, to $91.76$ I-AUROC and $94.99$ AUPRO at $K=2$ (from $94.09$ and $95.86$ at $K=1$), recovers at $K=16$ and then rises up to $K=64$ for image-level detection ($95.58$) and $K=128$ for AUPRO ($96.21$). Beyond that, image-level detection stays flat ($95.43$ at $K=128$, $95.16$ at $K=1024$), while AUPRO falls to $95.69$ at $K=1024$ as components become too sparse to estimate their subspaces reliably. The best $K$ for localization grows with the support data, from $K=128$ at 1- and 2-shot to $K=512$ at 4-shot (\cref{sec:supp_kshot}).

The dip at small $K$ is partly caused by comparing residuals of components with different dimensions (\cref{fig:k_dip}): 
a component with more available directions can reconstruct almost any patch well. With few, large components, the minimum in the scoring rule \eqref{eq:scoring} therefore favors the largest subspace instead of the best-matching region.
Giving all components the same number of directions $r_k$ reduces the image-level dip ($-0.96$ against $-2.33$ I-AUROC at $K=2$) but hardly the localization dip ($-0.68$ against $-0.87$ AUPRO). It also removes most of the gain at larger $K$ ($-0.05$ against $+1.50$ I-AUROC at $K=64$), and its AUPRO stays below the single subspace at every $K$, so the per-component $r_k$ of \eqref{eq:tau} is needed for the gain.

\textbf{Variance threshold $\boldsymbol{\tau}$.} The explained-variance threshold has little influence near the default: from $\tau=0.95$ to $0.995$ every metric stays within $0.32$ points of $\tau=0.99$. Performance drops slightly at $\tau=0.90$ ($-0.67$ I-AUROC, $-0.32$ AUPRO) and more at $\tau=0.999$ ($-1.44$ I-AUROC, $-0.54$ AUPRO; \cref{sec:supp_tau}). SubspaceAD \cite{lenderingSubspaceADTrainingFreeFewShot2026} likewise reports only small changes between $\tau=0.95$ and $0.99$ in the single subspace setting. 

\subsection{What drives the improvement}
\label{ssec:ablation}

\textbf{Per-cluster subspaces carry the gain.} Sharing one subspace across all components while keeping the cluster means (experiment \emph{shared covariance} in \cref{tab:ablation_paired}) falls back to SubspaceAD level on image level ($+0.09$ and $-0.01$ I-AUROC at 1/4 shots compared with SubspaceAD) and recovers only part of the AUPRO gap ($+0.15$ and $+0.11$ at 1/4 shots against LocalSubspaceAD's $+0.34$ and $+0.52$ gain over SubspaceAD). Centering every component on the global mean $\mu$ instead (experiment \emph{shared mean} in \cref{tab:ablation_paired}) shows consistent performance with LocalSubspaceAD at 1-shot (within $0.01$ points) and at 4-shot ($+0.18$ I-AUROC, $+0.10$ AUPRO). 

\textbf{Coherent clusters matter.} Assigning patches to components at random keeps the number of components but removes their relation to the data. Uniform random assignment (see \cref{tab:ablation_paired}) already improves I-AUROC over the baseline, while localization performance remains essentially unchanged ($+0.76$ and $+0.74$ I-AUROC, $-0.07$ and $+0.32$ AUPRO). Assigning patches at random to components with the same sizes as the k-means clusters shows that the size distribution of the clusters is not the source of the gains. Localization drops to SubspaceAD level or below (AUPRO $-0.33$ and $+0.03$) and image-level detection gains little ($+0.18$ and $+0.28$ I-AUROC).  Added capacity alone thus explains part of the gain, but the full gain in detection and localization requires components that follow the structure of the feature space. Plain k-means is sufficient to find them (\cref{fig:initialization}): random centers without iteration lose $2.95$/$2.65$ I-AUROC and $1.80$/$1.08$ AUPRO at 1/4 shots, and refining them by residual reassignment (VQPCA) does not recover the loss ($-2.92$/$-2.62$ I-AUROC) while the fit takes $45$\,min instead of $69$\,s at 4-shot.

\begin{table}[t]
\centering
\caption{Ablation on VisA: each variant compared to SubspaceAD, in points, mean $\pm$ std over seeds of the paired difference. DINOv3 rows are relative to SubspaceAD on DINOv3.}
\label{tab:ablation_paired}
\setlength{\tabcolsep}{2.5pt}
\footnotesize
\begin{tabular}{l@{\hspace{5pt}}cc@{\hspace{5pt}}cc}
\toprule
& \multicolumn{2}{c}{1-shot} & \multicolumn{2}{c}{4-shot} \\
\cmidrule(lr){2-3} \cmidrule(lr){4-5}
Method & I-AUROC & AUPRO & I-AUROC & AUPRO \\
\midrule
PatchCore (DINOv2-G) & -1.58{\tiny$\pm$0.68} & -1.59{\tiny$\pm$0.23} & -0.39{\tiny$\pm$0.48} & -1.14{\tiny$\pm$0.10} \\
LocalSubspaceAD ($K=128$) & +1.60{\tiny$\pm$0.25} & +0.34{\tiny$\pm$0.07} & +1.19{\tiny$\pm$0.17} & +0.52{\tiny$\pm$0.22} \\
\midrule
shared covariance & +0.09{\tiny$\pm$0.04} & +0.15{\tiny$\pm$0.05} & -0.01{\tiny$\pm$0.12} & +0.11{\tiny$\pm$0.01} \\
shared mean & +1.60{\tiny$\pm$0.19} & +0.34{\tiny$\pm$0.07} & +1.38{\tiny$\pm$0.26} & +0.63{\tiny$\pm$0.18} \\
\multicolumn{5}{l}{\textit{random assignment:}} \\
\quad k-means sizes & +0.18{\tiny$\pm$0.34} & -0.33{\tiny$\pm$0.06} & +0.28{\tiny$\pm$0.15} & +0.03{\tiny$\pm$0.02} \\
\quad uniform & +0.76{\tiny$\pm$0.37} & -0.07{\tiny$\pm$0.18} & +0.74{\tiny$\pm$0.09} & +0.32{\tiny$\pm$0.05} \\
\midrule
center bank k256, top 1\% & -1.16{\tiny$\pm$0.50} & -3.85{\tiny$\pm$0.30} & -1.05{\tiny$\pm$0.27} & -3.62{\tiny$\pm$0.25} \\
\midrule
\multicolumn{5}{l}{\textit{DINOv3}} \\
\quad LocalSubspaceAD K=1 & +2.21{\tiny$\pm$0.12} & +0.74{\tiny$\pm$0.13} & +1.41{\tiny$\pm$0.22} & +0.36{\tiny$\pm$0.03} \\
\quad LocalSubspaceAD K=128 & +3.02{\tiny$\pm$0.96} & +1.40{\tiny$\pm$0.10} & +2.84{\tiny$\pm$0.36} & +1.44{\tiny$\pm$0.12} \\
\bottomrule
\end{tabular}
\end{table}

\textbf{Not a memory bank.} If LocalSubspaceAD acted as a compressed memory bank, scoring by the distance to the nearest cluster center, $\min_k \|x_p - \mu_k\|^2$, would suffice. With $256$ centers this loses $1.05$ to $1.16$ I-AUROC and $3.62$ to $3.85$ AUPRO compared with SubspaceAD. PatchCore on the same features loses $0.39$ to $1.58$ I-AUROC and $1.14$ to $1.59$ AUPRO compared with SubspaceAD.

\section{Conclusion}
\label{sec:conclusion}

We proposed LocalSubspaceAD, a training-free few-shot anomaly detection method that models heterogeneous nominal data with a mixture of local subspaces fit to k-means clusters of foundation model features. On MVTec AD, VisA and Real-IAD it improves detection and localization over a single subspace (SubspaceAD) on both DINOv2 and DINOv3, with the largest gains on multi-instance categories and PCBs. The backbone of choice depends on the metric, for a single subspace as much as for local ones: DINOv2 features give the better image-level detection and DINOv3 features the better pixel-level AUPR. The gains from using LocalSubspaceAD are independent of this choice and hold on either backbone. The improvements stem from subspaces aligned to semantically coherent regions of feature space, not from cluster means or added capacity alone. On single-instance objects a single subspace remains slightly better in image-level detection, and the inference cost grows linearly with the number of components. Future work includes scoring only against the nearest components, the full-shot setting, and using the locality of the subspaces to detect logical anomalies.

\section{Acknowledgements}
This work was funded by the Austrian Research Promotion Agency (FFG) under the program ``Industrienahe Dissertationen'', grant no. FO999932323.

\section{LLM Usage Statement}
In writing this paper, Claude Opus 5.5 was used for editing, spell checking, grammar and readability enhancements. Further, Claude Opus 5.5 in Claude Code was used to generate and review code to produce experiments, figures and tables based on detailed specifications.
{
    \small
    \bibliographystyle{ieeenat_fullname}
    \bibliography{_mppcaad.bib}
}

\clearpage
\onecolumn 
\raggedbottom 
\setcounter{page}{1}
\begin{center}
    \Large
    \textbf{\thetitle}\\
    \vspace{0.5em}Supplementary Material\\
    \vspace{1.0em}
\end{center}

\suppfloatsetup

\renewcommand{\tablename}{Supp. Table}
\renewcommand{\figurename}{Supp. Figure}

\crefalias{section}{suppsection}
\crefalias{subsection}{suppsection}
\crefalias{subsubsection}{suppsection}
\crefalias{table}{supptable}
\crefalias{figure}{suppfigure}
\crefalias{equation}{suppequation}

\suppbibsetup

\FloatBarrier
\section{Full Few-Shot Results}
\label{sec:supp_fewshot}

\Cref{tab:fewshot_full} extends \cref{tab:fewshot} of the main paper to all five metrics and every shot count, on MVTec AD, VisA and Real-IAD.

\begin{table*}[t]
\centering
\caption{Few-shot anomaly detection, all metrics, mean $\pm$ std over 10 seeds. Best per dataset and shot count in \textbf{bold}, second \underline{underlined}. UniVAD~\citeS{guUniVADTrainingfreeUnified2025}
(official code, 448 px) uses CLIP ViT-L/14-336, DINOv2 ViT-g/14 without registers and DINO ViT-S/8 with Grounded-SAM component masks; UniVAD was not run on Real-IAD due to compute limitations.$^{\ddagger}$: As reported by the original authors; --: not reported. MVTec AD is shown here, VisA and Real-IAD on the next page in \cref{tab:fewshot_full_cont}.}
\label{tab:fewshot_full}
\footnotesize
\begin{tabular}{clccccc}
\toprule
Shots & Method & I-AUROC & I-AUPR & P-AUROC & P-AUPR & AUPRO \\
\midrule
\multicolumn{7}{c}{\textit{MVTec AD}} \\
\midrule
\multirow{10}{*}{1} & AnomalyDino+~\citeS{liFastRefFastPrototype2026}$^{\ddagger}$ & \underline{97.00} & -- & 96.50 & -- & -- \\
 & WinCLIP~\citeS{jeongWinCLIPZeroFewShot2023}$^{\ddagger}$ & 93.10{\scriptsize$\pm$2.00} & 96.50{\scriptsize$\pm$0.90} & -- & -- & -- \\
 & PatchCore~\citeS{rothTotalRecallIndustrial2022}$^{\ddagger}$ & 83.40{\scriptsize$\pm$0.60} & -- & 92.00{\scriptsize$\pm$0.20} & -- & 82.40{\scriptsize$\pm$0.30} \\
 & PromptAD~\citeS{liPromptADLearningPrompts2024}$^{\ddagger}$ & 94.60{\scriptsize$\pm$1.70} & -- & 95.90 & -- & -- \\
 & SubspaceAD (DINOv2-G) & 96.04{\scriptsize$\pm$1.38} & 98.05{\scriptsize$\pm$0.65} & 97.42{\scriptsize$\pm$0.12} & 58.26{\scriptsize$\pm$0.95} & 94.38{\scriptsize$\pm$0.16} \\
 & DuoAD (DINOv3-L) & 96.59{\scriptsize$\pm$0.86} & 98.23{\scriptsize$\pm$0.36} & 96.82{\scriptsize$\pm$0.20} & 63.63{\scriptsize$\pm$0.93} & 93.83{\scriptsize$\pm$0.32} \\
 & UniVAD (CLIP-L, DINOv2-g) & 96.06{\scriptsize$\pm$1.31} & 97.95{\scriptsize$\pm$0.76} & 96.12{\scriptsize$\pm$0.16} & 54.23{\scriptsize$\pm$0.75} & 92.64{\scriptsize$\pm$0.19} \\
 & SubspaceAD (DINOv3-L) & 94.97{\scriptsize$\pm$0.83} & 97.41{\scriptsize$\pm$0.38} & \underline{97.70}{\scriptsize$\pm$0.26} & \underline{68.45}{\scriptsize$\pm$0.89} & 93.37{\scriptsize$\pm$0.49} \\
 & \textbf{LocalSubspaceAD (DINOv3-L)} & 96.97{\scriptsize$\pm$0.99} & \underline{98.61}{\scriptsize$\pm$0.42} & \textbf{97.82}{\scriptsize$\pm$0.19} & \textbf{70.64}{\scriptsize$\pm$0.90} & \textbf{95.24}{\scriptsize$\pm$0.29} \\
 & \textbf{LocalSubspaceAD (DINOv2-G)} & \textbf{97.33}{\scriptsize$\pm$0.84} & \textbf{98.88}{\scriptsize$\pm$0.38} & 97.34{\scriptsize$\pm$0.13} & 58.72{\scriptsize$\pm$1.09} & \underline{94.77}{\scriptsize$\pm$0.20} \\
\cmidrule(lr){1-7}
\multirow{10}{*}{2} & AnomalyDino+~\citeS{liFastRefFastPrototype2026}$^{\ddagger}$ & 97.20 & -- & 96.70 & -- & -- \\
 & WinCLIP~\citeS{jeongWinCLIPZeroFewShot2023}$^{\ddagger}$ & 94.40{\scriptsize$\pm$1.30} & 97.00{\scriptsize$\pm$0.70} & -- & -- & -- \\
 & PatchCore~\citeS{rothTotalRecallIndustrial2022}$^{\ddagger}$ & 86.40{\scriptsize$\pm$0.90} & -- & 93.10{\scriptsize$\pm$0.20} & -- & 85.10{\scriptsize$\pm$0.30} \\
 & PromptAD~\citeS{liPromptADLearningPrompts2024}$^{\ddagger}$ & 95.70{\scriptsize$\pm$1.50} & -- & -- & -- & -- \\
 & SubspaceAD (DINOv2-G) & 96.76{\scriptsize$\pm$0.91} & 98.47{\scriptsize$\pm$0.36} & 97.73{\scriptsize$\pm$0.09} & 59.96{\scriptsize$\pm$0.71} & 94.83{\scriptsize$\pm$0.11} \\
 & DuoAD (DINOv3-L) & 97.20{\scriptsize$\pm$0.70} & 98.59{\scriptsize$\pm$0.27} & 97.15{\scriptsize$\pm$0.15} & 64.96{\scriptsize$\pm$0.84} & 94.45{\scriptsize$\pm$0.31} \\
 & UniVAD (CLIP-L, DINOv2-g) & 96.99{\scriptsize$\pm$1.08} & 98.49{\scriptsize$\pm$0.52} & 96.43{\scriptsize$\pm$0.12} & 55.50{\scriptsize$\pm$0.50} & 93.12{\scriptsize$\pm$0.15} \\
 & SubspaceAD (DINOv3-L) & 96.07{\scriptsize$\pm$0.48} & 97.94{\scriptsize$\pm$0.28} & \underline{97.95}{\scriptsize$\pm$0.11} & \underline{69.23}{\scriptsize$\pm$0.53} & 94.01{\scriptsize$\pm$0.32} \\
 & \textbf{LocalSubspaceAD (DINOv3-L)} & \underline{97.48}{\scriptsize$\pm$0.83} & \underline{98.88}{\scriptsize$\pm$0.33} & \textbf{98.04}{\scriptsize$\pm$0.13} & \textbf{71.06}{\scriptsize$\pm$0.83} & \textbf{95.72}{\scriptsize$\pm$0.28} \\
 & \textbf{LocalSubspaceAD (DINOv2-G)} & \textbf{98.00}{\scriptsize$\pm$0.44} & \textbf{99.20}{\scriptsize$\pm$0.12} & 97.72{\scriptsize$\pm$0.14} & 60.43{\scriptsize$\pm$0.53} & \underline{95.30}{\scriptsize$\pm$0.27} \\
\cmidrule(lr){1-7}
\multirow{9}{*}{4} & AnomalyDino+~\citeS{liFastRefFastPrototype2026}$^{\ddagger}$ & 97.40 & -- & 96.80 & -- & -- \\
 & WinCLIP~\citeS{jeongWinCLIPZeroFewShot2023}$^{\ddagger}$ & 95.20{\scriptsize$\pm$1.30} & 97.30{\scriptsize$\pm$0.60} & -- & -- & -- \\
 & PromptAD~\citeS{liPromptADLearningPrompts2024}$^{\ddagger}$ & 96.60{\scriptsize$\pm$0.90} & -- & 96.50 & -- & -- \\
 & SubspaceAD (DINOv2-G) & 97.66{\scriptsize$\pm$0.70} & 98.94{\scriptsize$\pm$0.28} & 97.91{\scriptsize$\pm$0.05} & 61.38{\scriptsize$\pm$0.82} & 95.16{\scriptsize$\pm$0.09} \\
 & DuoAD (DINOv3-L) & 98.01{\scriptsize$\pm$0.65} & 98.96{\scriptsize$\pm$0.29} & 97.41{\scriptsize$\pm$0.11} & 66.65{\scriptsize$\pm$0.88} & 95.07{\scriptsize$\pm$0.19} \\
 & UniVAD (CLIP-L, DINOv2-g) & 97.99{\scriptsize$\pm$0.89} & 98.99{\scriptsize$\pm$0.40} & 96.63{\scriptsize$\pm$0.07} & 56.55{\scriptsize$\pm$0.63} & 93.43{\scriptsize$\pm$0.12} \\
 & SubspaceAD (DINOv3-L) & 97.30{\scriptsize$\pm$0.65} & 98.54{\scriptsize$\pm$0.28} & \underline{98.07}{\scriptsize$\pm$0.09} & \underline{70.54}{\scriptsize$\pm$0.99} & 94.55{\scriptsize$\pm$0.25} \\
 & \textbf{LocalSubspaceAD (DINOv3-L)} & \underline{98.31}{\scriptsize$\pm$0.74} & \underline{99.29}{\scriptsize$\pm$0.27} & \textbf{98.26}{\scriptsize$\pm$0.10} & \textbf{71.99}{\scriptsize$\pm$0.91} & \textbf{96.26}{\scriptsize$\pm$0.15} \\
 & \textbf{LocalSubspaceAD (DINOv2-G)} & \textbf{98.58}{\scriptsize$\pm$0.44} & \textbf{99.47}{\scriptsize$\pm$0.15} & 97.96{\scriptsize$\pm$0.08} & 61.41{\scriptsize$\pm$0.91} & \underline{95.70}{\scriptsize$\pm$0.16} \\
\bottomrule
\end{tabular}
\par\vspace{2pt}
{\footnotesize $^{\ddagger}$\,Numbers as reported by the original authors, not rerun by us.}
\end{table*}

\begin{table*}[t]
\centering
\addtocounter{table}{-1}
\caption{Few-shot anomaly detection, all metrics, continued from \cref{tab:fewshot_full}: VisA and Real-IAD. Same setup and notation as \cref{tab:fewshot_full}.}
\label{tab:fewshot_full_cont}
\footnotesize
\begin{tabular}{clccccc}
\toprule
Shots & Method & I-AUROC & I-AUPR & P-AUROC & P-AUPR & AUPRO \\
\midrule
\multicolumn{7}{c}{\textit{VisA}} \\
\midrule
\multirow{9}{*}{1} & AnomalyDino+~\citeS{liFastRefFastPrototype2026}$^{\ddagger}$ & 84.60 & -- & 97.20 & -- & -- \\
 & WinCLIP~\citeS{jeongWinCLIPZeroFewShot2023}$^{\ddagger}$ & 83.80{\scriptsize$\pm$4.00} & 85.10{\scriptsize$\pm$4.00} & -- & -- & -- \\
 & PromptAD~\citeS{liPromptADLearningPrompts2024}$^{\ddagger}$ & 86.90{\scriptsize$\pm$2.30} & -- & -- & -- & -- \\
 & SubspaceAD (DINOv2-G) & 92.55{\scriptsize$\pm$0.79} & 92.25{\scriptsize$\pm$0.87} & 98.21{\scriptsize$\pm$0.06} & 36.96{\scriptsize$\pm$0.75} & \underline{95.40}{\scriptsize$\pm$0.11} \\
 & DuoAD (DINOv3-L) & 92.28{\scriptsize$\pm$0.64} & 92.45{\scriptsize$\pm$0.49} & 98.07{\scriptsize$\pm$0.08} & 45.21{\scriptsize$\pm$0.39} & 92.80{\scriptsize$\pm$0.26} \\
 & UniVAD (CLIP-L, DINOv2-g) & 93.16{\scriptsize$\pm$0.84} & \underline{93.82}{\scriptsize$\pm$0.60} & 97.98{\scriptsize$\pm$0.08} & 36.73{\scriptsize$\pm$0.58} & 92.64{\scriptsize$\pm$0.43} \\
 & SubspaceAD (DINOv3-L) & 90.36{\scriptsize$\pm$0.96} & 89.95{\scriptsize$\pm$0.76} & \underline{98.56}{\scriptsize$\pm$0.06} & \underline{50.17}{\scriptsize$\pm$0.85} & 93.42{\scriptsize$\pm$0.23} \\
 & \textbf{LocalSubspaceAD (DINOv3-L)} & \underline{93.57}{\scriptsize$\pm$0.58} & 93.62{\scriptsize$\pm$0.59} & \textbf{98.74}{\scriptsize$\pm$0.05} & \textbf{55.09}{\scriptsize$\pm$0.59} & 94.83{\scriptsize$\pm$0.18} \\
 & \textbf{LocalSubspaceAD (DINOv2-G)} & \textbf{94.00}{\scriptsize$\pm$0.74} & \textbf{94.38}{\scriptsize$\pm$0.76} & 98.38{\scriptsize$\pm$0.05} & 41.82{\scriptsize$\pm$0.81} & \textbf{95.71}{\scriptsize$\pm$0.12} \\
\cmidrule(lr){1-7}
\multirow{9}{*}{2} & AnomalyDino+~\citeS{liFastRefFastPrototype2026}$^{\ddagger}$ & 84.80 & -- & 97.30 & -- & -- \\
 & WinCLIP~\citeS{jeongWinCLIPZeroFewShot2023}$^{\ddagger}$ & 84.60{\scriptsize$\pm$2.40} & 85.80{\scriptsize$\pm$2.70} & -- & -- & -- \\
 & PromptAD~\citeS{liPromptADLearningPrompts2024}$^{\ddagger}$ & 88.30{\scriptsize$\pm$2.00} & -- & -- & -- & -- \\
 & SubspaceAD (DINOv2-G) & 93.99{\scriptsize$\pm$0.59} & 93.73{\scriptsize$\pm$0.68} & 98.34{\scriptsize$\pm$0.03} & 39.39{\scriptsize$\pm$0.83} & \underline{95.77}{\scriptsize$\pm$0.08} \\
 & DuoAD (DINOv3-L) & 93.10{\scriptsize$\pm$0.61} & 92.99{\scriptsize$\pm$0.46} & 98.35{\scriptsize$\pm$0.07} & 47.50{\scriptsize$\pm$0.50} & 93.80{\scriptsize$\pm$0.17} \\
 & UniVAD (CLIP-L, DINOv2-g) & 94.77{\scriptsize$\pm$0.53} & \underline{95.26}{\scriptsize$\pm$0.42} & 98.24{\scriptsize$\pm$0.02} & 39.24{\scriptsize$\pm$0.45} & 93.40{\scriptsize$\pm$0.25} \\
 & SubspaceAD (DINOv3-L) & 91.84{\scriptsize$\pm$0.58} & 91.18{\scriptsize$\pm$0.56} & \underline{98.65}{\scriptsize$\pm$0.04} & \underline{51.67}{\scriptsize$\pm$0.68} & 94.00{\scriptsize$\pm$0.11} \\
 & \textbf{LocalSubspaceAD (DINOv3-L)} & \underline{94.80}{\scriptsize$\pm$0.54} & 94.43{\scriptsize$\pm$0.48} & \textbf{98.87}{\scriptsize$\pm$0.03} & \textbf{55.53}{\scriptsize$\pm$0.42} & 95.46{\scriptsize$\pm$0.09} \\
 & \textbf{LocalSubspaceAD (DINOv2-G)} & \textbf{95.20}{\scriptsize$\pm$0.53} & \textbf{95.36}{\scriptsize$\pm$0.62} & 98.59{\scriptsize$\pm$0.03} & 43.76{\scriptsize$\pm$0.79} & \textbf{96.20}{\scriptsize$\pm$0.12} \\
\cmidrule(lr){1-7}
\multirow{9}{*}{4} & AnomalyDino+~\citeS{liFastRefFastPrototype2026}$^{\ddagger}$ & 87.10 & -- & 97.90 & -- & -- \\
 & WinCLIP~\citeS{jeongWinCLIPZeroFewShot2023}$^{\ddagger}$ & 87.30{\scriptsize$\pm$1.80} & 84.20{\scriptsize$\pm$1.60} & -- & -- & -- \\
 & PromptAD~\citeS{liPromptADLearningPrompts2024}$^{\ddagger}$ & 89.10{\scriptsize$\pm$1.70} & -- & -- & -- & -- \\
 & SubspaceAD (DINOv2-G) & 94.79{\scriptsize$\pm$0.43} & 94.50{\scriptsize$\pm$0.62} & 98.44{\scriptsize$\pm$0.03} & 40.66{\scriptsize$\pm$0.47} & \underline{96.00}{\scriptsize$\pm$0.12} \\
 & DuoAD (DINOv3-L) & 93.82{\scriptsize$\pm$0.35} & 93.57{\scriptsize$\pm$0.27} & 98.55{\scriptsize$\pm$0.03} & 49.03{\scriptsize$\pm$0.46} & 94.55{\scriptsize$\pm$0.11} \\
 & UniVAD (CLIP-L, DINOv2-g) & 95.75{\scriptsize$\pm$0.42} & \underline{96.00}{\scriptsize$\pm$0.38} & 98.39{\scriptsize$\pm$0.03} & 40.66{\scriptsize$\pm$0.49} & 93.78{\scriptsize$\pm$0.10} \\
 & SubspaceAD (DINOv3-L) & 93.11{\scriptsize$\pm$0.43} & 92.23{\scriptsize$\pm$0.41} & 98.71{\scriptsize$\pm$0.03} & \underline{52.47}{\scriptsize$\pm$0.62} & 94.31{\scriptsize$\pm$0.09} \\
 & \textbf{LocalSubspaceAD (DINOv3-L)} & \underline{95.91}{\scriptsize$\pm$0.51} & 95.27{\scriptsize$\pm$0.45} & \textbf{98.99}{\scriptsize$\pm$0.02} & \textbf{54.94}{\scriptsize$\pm$0.39} & 95.69{\scriptsize$\pm$0.12} \\
 & \textbf{LocalSubspaceAD (DINOv2-G)} & \textbf{96.21}{\scriptsize$\pm$0.39} & \textbf{96.05}{\scriptsize$\pm$0.41} & \underline{98.78}{\scriptsize$\pm$0.04} & 45.00{\scriptsize$\pm$0.96} & \textbf{96.55}{\scriptsize$\pm$0.11} \\
\midrule
\multicolumn{7}{c}{\textit{Real-IAD}} \\
\midrule
\multirow{5}{*}{1} & SubspaceAD (DINOv2-G) & \underline{88.60}{\scriptsize$\pm$0.50} & \underline{86.77}{\scriptsize$\pm$0.47} & \underline{99.01}{\scriptsize$\pm$0.06} & 44.02{\scriptsize$\pm$0.90} & \underline{97.40}{\scriptsize$\pm$0.21} \\
 & DuoAD (DINOv3-L) & 88.18{\scriptsize$\pm$0.67} & 85.79{\scriptsize$\pm$0.69} & 98.71{\scriptsize$\pm$0.04} & 50.96{\scriptsize$\pm$0.85} & 96.25{\scriptsize$\pm$0.19} \\
 & SubspaceAD (DINOv3-L) & 83.62{\scriptsize$\pm$0.98} & 81.53{\scriptsize$\pm$0.84} & 98.35{\scriptsize$\pm$0.09} & \underline{51.14}{\scriptsize$\pm$1.08} & 94.15{\scriptsize$\pm$0.34} \\
 & \textbf{LocalSubspaceAD (DINOv3-L)} & 88.20{\scriptsize$\pm$0.52} & 86.19{\scriptsize$\pm$0.47} & 98.74{\scriptsize$\pm$0.05} & \textbf{56.70}{\scriptsize$\pm$0.75} & 96.26{\scriptsize$\pm$0.19} \\
 & \textbf{LocalSubspaceAD (DINOv2-G)} & \textbf{90.88}{\scriptsize$\pm$0.36} & \textbf{88.93}{\scriptsize$\pm$0.70} & \textbf{99.19}{\scriptsize$\pm$0.04} & 48.72{\scriptsize$\pm$0.85} & \textbf{97.77}{\scriptsize$\pm$0.16} \\
\cmidrule(lr){1-7}
\multirow{5}{*}{2} & SubspaceAD (DINOv2-G) & \underline{90.30}{\scriptsize$\pm$0.42} & \underline{88.43}{\scriptsize$\pm$0.49} & \underline{99.10}{\scriptsize$\pm$0.03} & 45.67{\scriptsize$\pm$0.59} & \underline{97.69}{\scriptsize$\pm$0.09} \\
 & DuoAD (DINOv3-L) & 89.85{\scriptsize$\pm$0.51} & 87.26{\scriptsize$\pm$0.56} & 98.86{\scriptsize$\pm$0.02} & 52.41{\scriptsize$\pm$0.52} & 96.73{\scriptsize$\pm$0.07} \\
 & SubspaceAD (DINOv3-L) & 86.04{\scriptsize$\pm$0.69} & 83.87{\scriptsize$\pm$0.80} & 98.56{\scriptsize$\pm$0.04} & \underline{52.88}{\scriptsize$\pm$0.80} & 95.13{\scriptsize$\pm$0.15} \\
 & \textbf{LocalSubspaceAD (DINOv3-L)} & 89.91{\scriptsize$\pm$0.41} & 87.65{\scriptsize$\pm$0.52} & 98.86{\scriptsize$\pm$0.03} & \textbf{57.33}{\scriptsize$\pm$0.50} & 96.77{\scriptsize$\pm$0.11} \\
 & \textbf{LocalSubspaceAD (DINOv2-G)} & \textbf{92.09}{\scriptsize$\pm$0.34} & \textbf{90.04}{\scriptsize$\pm$0.70} & \textbf{99.29}{\scriptsize$\pm$0.03} & 49.95{\scriptsize$\pm$0.59} & \textbf{98.01}{\scriptsize$\pm$0.07} \\
\cmidrule(lr){1-7}
\multirow{5}{*}{4} & SubspaceAD (DINOv2-G) & \underline{91.65}{\scriptsize$\pm$0.40} & \underline{90.01}{\scriptsize$\pm$0.53} & \underline{99.18}{\scriptsize$\pm$0.01} & 47.44{\scriptsize$\pm$0.30} & \underline{97.96}{\scriptsize$\pm$0.06} \\
 & DuoAD (DINOv3-L) & 91.17{\scriptsize$\pm$0.35} & 88.67{\scriptsize$\pm$0.49} & 98.96{\scriptsize$\pm$0.02} & 53.86{\scriptsize$\pm$0.41} & 97.08{\scriptsize$\pm$0.05} \\
 & SubspaceAD (DINOv3-L) & 87.69{\scriptsize$\pm$0.46} & 85.71{\scriptsize$\pm$0.63} & 98.69{\scriptsize$\pm$0.02} & \underline{54.25}{\scriptsize$\pm$0.56} & 95.72{\scriptsize$\pm$0.09} \\
 & \textbf{LocalSubspaceAD (DINOv3-L)} & 90.99{\scriptsize$\pm$0.25} & 88.95{\scriptsize$\pm$0.43} & 98.98{\scriptsize$\pm$0.01} & \textbf{57.36}{\scriptsize$\pm$0.35} & 97.12{\scriptsize$\pm$0.08} \\
 & \textbf{LocalSubspaceAD (DINOv2-G)} & \textbf{93.03}{\scriptsize$\pm$0.38} & \textbf{91.31}{\scriptsize$\pm$0.48} & \textbf{99.39}{\scriptsize$\pm$0.02} & 50.65{\scriptsize$\pm$0.19} & \textbf{98.23}{\scriptsize$\pm$0.12} \\
\bottomrule
\end{tabular}
\par\vspace{2pt}
{\footnotesize $^{\ddagger}$\,Numbers as reported by the original authors, not rerun by us.}
\end{table*}

\FloatBarrier
\section{Paired Differences}
\label{sec:supp_paired}
The variation in performance caused by the choice of support set is often as large as the performance differences between methods. We therefore also compare methods by paired differences (\cref{tab:fewshot_summary}): per seed, the difference of the two methods' category means on the same support sets, reported as mean $\pm$ std over seeds. A difference with the same sign at every seed and shot count is set in bold. \Cref{tab:fewshot_paired} splits the paired differences of \cref{tab:fewshot_summary} by shot count, and \cref{tab:fewshot_paired_all} reports them on all metrics against every other method.

\begin{table*}[t]
\centering
\caption{LocalSubspaceAD minus the baselines per shot count, in points: mean $\pm$ std over 10 seeds of the paired difference. UniVAD~\citeS{guUniVADTrainingfreeUnified2025}
is compared against both backbones and was not run on Real-IAD (--). \textbf{Bold}: the same sign at every seed.}
\label{tab:fewshot_paired}
\setlength{\tabcolsep}{2.5pt}
\footnotesize
\begin{tabular}{cll@{\hspace{5pt}}ccc@{\hspace{5pt}}ccc@{\hspace{5pt}}ccc}
\toprule
& & & \multicolumn{3}{c}{MVTec AD} & \multicolumn{3}{c}{VisA} & \multicolumn{3}{c}{Real-IAD} \\
\cmidrule(lr){4-6} \cmidrule(lr){7-9} \cmidrule(lr){10-12}
Shots & Backbone & LocalSubspaceAD minus & I-AUROC & P-AUPR & AUPRO & I-AUROC & P-AUPR & AUPRO & I-AUROC & P-AUPR & AUPRO \\
\midrule
\multirow{5}{*}{1} & \multirow{2}{*}{DINOv2} & SubspaceAD & \textbf{+1.30}{\tiny$\pm$0.70} & +0.46{\tiny$\pm$0.69} & \textbf{+0.39}{\tiny$\pm$0.13} & \textbf{+1.45}{\tiny$\pm$0.73} & \textbf{+4.86}{\tiny$\pm$0.99} & \textbf{+0.32}{\tiny$\pm$0.12} & \textbf{+2.28}{\tiny$\pm$0.40} & \textbf{+4.70}{\tiny$\pm$0.70} & \textbf{+0.37}{\tiny$\pm$0.11} \\
 &  & UniVAD & \textbf{+1.28}{\tiny$\pm$0.71} & \textbf{+4.49}{\tiny$\pm$0.70} & \textbf{+2.13}{\tiny$\pm$0.21} & \textbf{+0.84}{\tiny$\pm$0.46} & \textbf{+5.09}{\tiny$\pm$0.75} & \textbf{+3.07}{\tiny$\pm$0.40} & -- & -- & -- \\
\addlinespace[3pt]
 & \multirow{3}{*}{DINOv3} & SubspaceAD & \textbf{+2.00}{\tiny$\pm$0.43} & \textbf{+2.20}{\tiny$\pm$0.47} & \textbf{+1.87}{\tiny$\pm$0.29} & \textbf{+3.22}{\tiny$\pm$0.94} & \textbf{+4.93}{\tiny$\pm$0.73} & \textbf{+1.42}{\tiny$\pm$0.21} & \textbf{+4.57}{\tiny$\pm$0.53} & \textbf{+5.56}{\tiny$\pm$0.59} & \textbf{+2.10}{\tiny$\pm$0.21} \\
 &  & DuoAD & \textbf{+0.38}{\tiny$\pm$0.34} & \textbf{+7.01}{\tiny$\pm$0.34} & \textbf{+1.41}{\tiny$\pm$0.15} & \textbf{+1.29}{\tiny$\pm$0.54} & \textbf{+9.88}{\tiny$\pm$0.61} & \textbf{+2.03}{\tiny$\pm$0.17} & +0.02{\tiny$\pm$0.39} & \textbf{+5.73}{\tiny$\pm$0.47} & +0.01{\tiny$\pm$0.10} \\
 &  & UniVAD & \textbf{+0.92}{\tiny$\pm$0.54} & \textbf{+16.41}{\tiny$\pm$0.39} & \textbf{+2.60}{\tiny$\pm$0.22} & +0.42{\tiny$\pm$0.85} & \textbf{+18.36}{\tiny$\pm$0.88} & \textbf{+2.20}{\tiny$\pm$0.46} & -- & -- & -- \\
\cmidrule(lr){1-12}
\multirow{5}{*}{2} & \multirow{2}{*}{DINOv2} & SubspaceAD & \textbf{+1.24}{\tiny$\pm$0.58} & +0.47{\tiny$\pm$0.77} & \textbf{+0.47}{\tiny$\pm$0.16} & \textbf{+1.21}{\tiny$\pm$0.44} & \textbf{+4.37}{\tiny$\pm$0.95} & \textbf{+0.44}{\tiny$\pm$0.13} & \textbf{+1.79}{\tiny$\pm$0.32} & \textbf{+4.27}{\tiny$\pm$0.51} & \textbf{+0.32}{\tiny$\pm$0.06} \\
 &  & UniVAD & +1.01{\tiny$\pm$0.76} & \textbf{+4.93}{\tiny$\pm$0.55} & \textbf{+2.17}{\tiny$\pm$0.18} & +0.42{\tiny$\pm$0.48} & \textbf{+4.52}{\tiny$\pm$0.91} & \textbf{+2.80}{\tiny$\pm$0.19} & -- & -- & -- \\
\addlinespace[3pt]
 & \multirow{3}{*}{DINOv3} & SubspaceAD & \textbf{+1.41}{\tiny$\pm$0.57} & \textbf{+1.83}{\tiny$\pm$0.50} & \textbf{+1.71}{\tiny$\pm$0.16} & \textbf{+2.95}{\tiny$\pm$0.30} & \textbf{+3.87}{\tiny$\pm$0.62} & \textbf{+1.46}{\tiny$\pm$0.13} & \textbf{+3.87}{\tiny$\pm$0.48} & \textbf{+4.44}{\tiny$\pm$0.60} & \textbf{+1.64}{\tiny$\pm$0.11} \\
 &  & DuoAD & +0.27{\tiny$\pm$0.22} & \textbf{+6.10}{\tiny$\pm$0.33} & \textbf{+1.27}{\tiny$\pm$0.15} & \textbf{+1.70}{\tiny$\pm$0.52} & \textbf{+8.03}{\tiny$\pm$0.47} & \textbf{+1.66}{\tiny$\pm$0.16} & +0.06{\tiny$\pm$0.31} & \textbf{+4.92}{\tiny$\pm$0.43} & +0.04{\tiny$\pm$0.08} \\
 &  & UniVAD & +0.48{\tiny$\pm$0.37} & \textbf{+15.56}{\tiny$\pm$0.52} & \textbf{+2.59}{\tiny$\pm$0.16} & +0.02{\tiny$\pm$0.55} & \textbf{+16.30}{\tiny$\pm$0.72} & \textbf{+2.06}{\tiny$\pm$0.20} & -- & -- & -- \\
\cmidrule(lr){1-12}
\multirow{5}{*}{4} & \multirow{2}{*}{DINOv2} & SubspaceAD & \textbf{+0.92}{\tiny$\pm$0.43} & +0.03{\tiny$\pm$0.57} & \textbf{+0.54}{\tiny$\pm$0.10} & \textbf{+1.42}{\tiny$\pm$0.52} & \textbf{+4.34}{\tiny$\pm$1.10} & \textbf{+0.55}{\tiny$\pm$0.16} & \textbf{+1.38}{\tiny$\pm$0.11} & \textbf{+3.21}{\tiny$\pm$0.39} & \textbf{+0.27}{\tiny$\pm$0.08} \\
 &  & UniVAD & +0.59{\tiny$\pm$0.60} & \textbf{+4.86}{\tiny$\pm$0.80} & \textbf{+2.28}{\tiny$\pm$0.13} & +0.46{\tiny$\pm$0.55} & \textbf{+4.34}{\tiny$\pm$0.99} & \textbf{+2.77}{\tiny$\pm$0.07} & -- & -- & -- \\
\addlinespace[3pt]
 & \multirow{3}{*}{DINOv3} & SubspaceAD & \textbf{+1.02}{\tiny$\pm$0.34} & \textbf{+1.45}{\tiny$\pm$0.45} & \textbf{+1.71}{\tiny$\pm$0.14} & \textbf{+2.79}{\tiny$\pm$0.36} & \textbf{+2.47}{\tiny$\pm$0.55} & \textbf{+1.38}{\tiny$\pm$0.15} & \textbf{+3.30}{\tiny$\pm$0.30} & \textbf{+3.11}{\tiny$\pm$0.45} & \textbf{+1.40}{\tiny$\pm$0.08} \\
 &  & DuoAD & \textbf{+0.30}{\tiny$\pm$0.23} & \textbf{+5.34}{\tiny$\pm$0.26} & \textbf{+1.19}{\tiny$\pm$0.08} & \textbf{+2.09}{\tiny$\pm$0.46} & \textbf{+5.91}{\tiny$\pm$0.37} & \textbf{+1.14}{\tiny$\pm$0.14} & -0.19{\tiny$\pm$0.20} & \textbf{+3.50}{\tiny$\pm$0.45} & +0.04{\tiny$\pm$0.06} \\
 &  & UniVAD & +0.32{\tiny$\pm$0.42} & \textbf{+15.44}{\tiny$\pm$0.68} & \textbf{+2.83}{\tiny$\pm$0.10} & +0.16{\tiny$\pm$0.60} & \textbf{+14.28}{\tiny$\pm$0.52} & \textbf{+1.91}{\tiny$\pm$0.14} & -- & -- & -- \\
\bottomrule
\end{tabular}
\end{table*}

\begin{table*}[t]
\centering
\caption{LocalSubspaceAD minus each other method on all metrics, in points: mean $\pm$ std over 10 seeds of the paired difference. \textbf{Bold}: the same sign at every seed. UniVAD~\citeS{guUniVADTrainingfreeUnified2025}
was not run on Real-IAD.}
\label{tab:fewshot_paired_all}
\footnotesize
\begin{tabular}{cllccccc}
\toprule
Shots & Backbone & LocalSubspaceAD minus & I-AUROC & I-AUPR & P-AUROC & P-AUPR & AUPRO \\
\midrule
\multicolumn{8}{c}{\textit{MVTec AD}} \\
\midrule
\multirow{6}{*}{1} & DINOv2 & SubspaceAD & \textbf{+1.30}{\scriptsize$\pm$0.70} & \textbf{+0.84}{\scriptsize$\pm$0.37} & \textbf{-0.08}{\scriptsize$\pm$0.06} & +0.46{\scriptsize$\pm$0.69} & \textbf{+0.39}{\scriptsize$\pm$0.13} \\
 &  & DuoAD & \textbf{+0.74}{\scriptsize$\pm$0.34} & \textbf{+0.65}{\scriptsize$\pm$0.14} & \textbf{+0.52}{\scriptsize$\pm$0.09} & \textbf{-4.91}{\scriptsize$\pm$0.77} & \textbf{+0.94}{\scriptsize$\pm$0.19} \\
 &  & UniVAD & \textbf{+1.28}{\scriptsize$\pm$0.71} & +0.93{\scriptsize$\pm$0.56} & \textbf{+1.22}{\scriptsize$\pm$0.08} & \textbf{+4.49}{\scriptsize$\pm$0.70} & \textbf{+2.13}{\scriptsize$\pm$0.21} \\
 & DINOv3 & SubspaceAD & \textbf{+2.00}{\scriptsize$\pm$0.43} & \textbf{+1.20}{\scriptsize$\pm$0.27} & \textbf{+0.12}{\scriptsize$\pm$0.10} & \textbf{+2.20}{\scriptsize$\pm$0.47} & \textbf{+1.87}{\scriptsize$\pm$0.29} \\
 &  & DuoAD & \textbf{+0.38}{\scriptsize$\pm$0.34} & \textbf{+0.37}{\scriptsize$\pm$0.19} & \textbf{+1.00}{\scriptsize$\pm$0.04} & \textbf{+7.01}{\scriptsize$\pm$0.34} & \textbf{+1.41}{\scriptsize$\pm$0.15} \\
 &  & UniVAD & \textbf{+0.92}{\scriptsize$\pm$0.54} & +0.66{\scriptsize$\pm$0.55} & \textbf{+1.70}{\scriptsize$\pm$0.09} & \textbf{+16.41}{\scriptsize$\pm$0.39} & \textbf{+2.60}{\scriptsize$\pm$0.22} \\
\cmidrule(lr){1-8}
\multirow{6}{*}{2} & DINOv2 & SubspaceAD & \textbf{+1.24}{\scriptsize$\pm$0.58} & \textbf{+0.72}{\scriptsize$\pm$0.33} & -0.01{\scriptsize$\pm$0.06} & +0.47{\scriptsize$\pm$0.77} & \textbf{+0.47}{\scriptsize$\pm$0.16} \\
 &  & DuoAD & \textbf{+0.80}{\scriptsize$\pm$0.36} & \textbf{+0.60}{\scriptsize$\pm$0.21} & \textbf{+0.56}{\scriptsize$\pm$0.06} & \textbf{-4.53}{\scriptsize$\pm$0.72} & \textbf{+0.85}{\scriptsize$\pm$0.16} \\
 &  & UniVAD & +1.01{\scriptsize$\pm$0.76} & \textbf{+0.71}{\scriptsize$\pm$0.49} & \textbf{+1.29}{\scriptsize$\pm$0.08} & \textbf{+4.93}{\scriptsize$\pm$0.55} & \textbf{+2.17}{\scriptsize$\pm$0.18} \\
 & DINOv3 & SubspaceAD & \textbf{+1.41}{\scriptsize$\pm$0.57} & \textbf{+0.93}{\scriptsize$\pm$0.28} & \textbf{+0.09}{\scriptsize$\pm$0.07} & \textbf{+1.83}{\scriptsize$\pm$0.50} & \textbf{+1.71}{\scriptsize$\pm$0.16} \\
 &  & DuoAD & +0.27{\scriptsize$\pm$0.22} & \textbf{+0.28}{\scriptsize$\pm$0.14} & \textbf{+0.88}{\scriptsize$\pm$0.06} & \textbf{+6.10}{\scriptsize$\pm$0.33} & \textbf{+1.27}{\scriptsize$\pm$0.15} \\
 &  & UniVAD & +0.48{\scriptsize$\pm$0.37} & +0.38{\scriptsize$\pm$0.31} & \textbf{+1.61}{\scriptsize$\pm$0.08} & \textbf{+15.56}{\scriptsize$\pm$0.52} & \textbf{+2.59}{\scriptsize$\pm$0.16} \\
\cmidrule(lr){1-8}
\multirow{6}{*}{4} & DINOv2 & SubspaceAD & \textbf{+0.92}{\scriptsize$\pm$0.43} & \textbf{+0.53}{\scriptsize$\pm$0.23} & +0.05{\scriptsize$\pm$0.05} & +0.03{\scriptsize$\pm$0.57} & \textbf{+0.54}{\scriptsize$\pm$0.10} \\
 &  & DuoAD & \textbf{+0.57}{\scriptsize$\pm$0.31} & \textbf{+0.50}{\scriptsize$\pm$0.17} & \textbf{+0.55}{\scriptsize$\pm$0.05} & \textbf{-5.24}{\scriptsize$\pm$0.50} & \textbf{+0.63}{\scriptsize$\pm$0.10} \\
 &  & UniVAD & +0.59{\scriptsize$\pm$0.60} & \textbf{+0.48}{\scriptsize$\pm$0.31} & \textbf{+1.33}{\scriptsize$\pm$0.08} & \textbf{+4.86}{\scriptsize$\pm$0.80} & \textbf{+2.28}{\scriptsize$\pm$0.13} \\
 & DINOv3 & SubspaceAD & \textbf{+1.02}{\scriptsize$\pm$0.34} & \textbf{+0.75}{\scriptsize$\pm$0.20} & \textbf{+0.19}{\scriptsize$\pm$0.03} & \textbf{+1.45}{\scriptsize$\pm$0.45} & \textbf{+1.71}{\scriptsize$\pm$0.14} \\
 &  & DuoAD & \textbf{+0.30}{\scriptsize$\pm$0.23} & \textbf{+0.32}{\scriptsize$\pm$0.16} & \textbf{+0.85}{\scriptsize$\pm$0.03} & \textbf{+5.34}{\scriptsize$\pm$0.26} & \textbf{+1.19}{\scriptsize$\pm$0.08} \\
 &  & UniVAD & +0.32{\scriptsize$\pm$0.42} & +0.30{\scriptsize$\pm$0.21} & \textbf{+1.63}{\scriptsize$\pm$0.08} & \textbf{+15.44}{\scriptsize$\pm$0.68} & \textbf{+2.83}{\scriptsize$\pm$0.10} \\
\midrule
\multicolumn{8}{c}{\textit{VisA}} \\
\midrule
\multirow{6}{*}{1} & DINOv2 & SubspaceAD & \textbf{+1.45}{\scriptsize$\pm$0.73} & \textbf{+2.13}{\scriptsize$\pm$0.86} & \textbf{+0.17}{\scriptsize$\pm$0.04} & \textbf{+4.86}{\scriptsize$\pm$0.99} & \textbf{+0.32}{\scriptsize$\pm$0.12} \\
 &  & DuoAD & \textbf{+1.71}{\scriptsize$\pm$0.52} & \textbf{+1.93}{\scriptsize$\pm$0.46} & \textbf{+0.32}{\scriptsize$\pm$0.07} & \textbf{-3.40}{\scriptsize$\pm$0.90} & \textbf{+2.91}{\scriptsize$\pm$0.27} \\
 &  & UniVAD & \textbf{+0.84}{\scriptsize$\pm$0.46} & +0.55{\scriptsize$\pm$0.51} & \textbf{+0.40}{\scriptsize$\pm$0.09} & \textbf{+5.09}{\scriptsize$\pm$0.75} & \textbf{+3.07}{\scriptsize$\pm$0.40} \\
 & DINOv3 & SubspaceAD & \textbf{+3.22}{\scriptsize$\pm$0.94} & \textbf{+3.67}{\scriptsize$\pm$0.90} & \textbf{+0.19}{\scriptsize$\pm$0.04} & \textbf{+4.93}{\scriptsize$\pm$0.73} & \textbf{+1.42}{\scriptsize$\pm$0.21} \\
 &  & DuoAD & \textbf{+1.29}{\scriptsize$\pm$0.54} & \textbf{+1.17}{\scriptsize$\pm$0.59} & \textbf{+0.68}{\scriptsize$\pm$0.05} & \textbf{+9.88}{\scriptsize$\pm$0.61} & \textbf{+2.03}{\scriptsize$\pm$0.17} \\
 &  & UniVAD & +0.42{\scriptsize$\pm$0.85} & -0.20{\scriptsize$\pm$0.66} & \textbf{+0.77}{\scriptsize$\pm$0.09} & \textbf{+18.36}{\scriptsize$\pm$0.88} & \textbf{+2.20}{\scriptsize$\pm$0.46} \\
\cmidrule(lr){1-8}
\multirow{6}{*}{2} & DINOv2 & SubspaceAD & \textbf{+1.21}{\scriptsize$\pm$0.44} & \textbf{+1.63}{\scriptsize$\pm$0.48} & \textbf{+0.25}{\scriptsize$\pm$0.03} & \textbf{+4.37}{\scriptsize$\pm$0.95} & \textbf{+0.44}{\scriptsize$\pm$0.13} \\
 &  & DuoAD & \textbf{+2.10}{\scriptsize$\pm$0.59} & \textbf{+2.37}{\scriptsize$\pm$0.60} & \textbf{+0.24}{\scriptsize$\pm$0.06} & \textbf{-3.74}{\scriptsize$\pm$0.82} & \textbf{+2.40}{\scriptsize$\pm$0.18} \\
 &  & UniVAD & +0.42{\scriptsize$\pm$0.48} & +0.10{\scriptsize$\pm$0.65} & \textbf{+0.35}{\scriptsize$\pm$0.03} & \textbf{+4.52}{\scriptsize$\pm$0.91} & \textbf{+2.80}{\scriptsize$\pm$0.19} \\
 & DINOv3 & SubspaceAD & \textbf{+2.95}{\scriptsize$\pm$0.30} & \textbf{+3.25}{\scriptsize$\pm$0.32} & \textbf{+0.22}{\scriptsize$\pm$0.02} & \textbf{+3.87}{\scriptsize$\pm$0.62} & \textbf{+1.46}{\scriptsize$\pm$0.13} \\
 &  & DuoAD & \textbf{+1.70}{\scriptsize$\pm$0.52} & \textbf{+1.44}{\scriptsize$\pm$0.40} & \textbf{+0.52}{\scriptsize$\pm$0.05} & \textbf{+8.03}{\scriptsize$\pm$0.47} & \textbf{+1.66}{\scriptsize$\pm$0.16} \\
 &  & UniVAD & +0.02{\scriptsize$\pm$0.55} & \textbf{-0.83}{\scriptsize$\pm$0.45} & \textbf{+0.63}{\scriptsize$\pm$0.03} & \textbf{+16.30}{\scriptsize$\pm$0.72} & \textbf{+2.06}{\scriptsize$\pm$0.20} \\
\cmidrule(lr){1-8}
\multirow{6}{*}{4} & DINOv2 & SubspaceAD & \textbf{+1.42}{\scriptsize$\pm$0.52} & \textbf{+1.56}{\scriptsize$\pm$0.63} & \textbf{+0.34}{\scriptsize$\pm$0.05} & \textbf{+4.34}{\scriptsize$\pm$1.10} & \textbf{+0.55}{\scriptsize$\pm$0.16} \\
 &  & DuoAD & \textbf{+2.39}{\scriptsize$\pm$0.35} & \textbf{+2.49}{\scriptsize$\pm$0.35} & \textbf{+0.23}{\scriptsize$\pm$0.03} & \textbf{-4.03}{\scriptsize$\pm$0.88} & \textbf{+2.00}{\scriptsize$\pm$0.17} \\
 &  & UniVAD & +0.46{\scriptsize$\pm$0.55} & +0.05{\scriptsize$\pm$0.56} & \textbf{+0.39}{\scriptsize$\pm$0.04} & \textbf{+4.34}{\scriptsize$\pm$0.99} & \textbf{+2.77}{\scriptsize$\pm$0.07} \\
 & DINOv3 & SubspaceAD & \textbf{+2.79}{\scriptsize$\pm$0.36} & \textbf{+3.04}{\scriptsize$\pm$0.37} & \textbf{+0.28}{\scriptsize$\pm$0.03} & \textbf{+2.47}{\scriptsize$\pm$0.55} & \textbf{+1.38}{\scriptsize$\pm$0.15} \\
 &  & DuoAD & \textbf{+2.09}{\scriptsize$\pm$0.46} & \textbf{+1.70}{\scriptsize$\pm$0.49} & \textbf{+0.44}{\scriptsize$\pm$0.03} & \textbf{+5.91}{\scriptsize$\pm$0.37} & \textbf{+1.14}{\scriptsize$\pm$0.14} \\
 &  & UniVAD & +0.16{\scriptsize$\pm$0.60} & -0.73{\scriptsize$\pm$0.54} & \textbf{+0.61}{\scriptsize$\pm$0.02} & \textbf{+14.28}{\scriptsize$\pm$0.52} & \textbf{+1.91}{\scriptsize$\pm$0.14} \\
\midrule
\multicolumn{8}{c}{\textit{Real-IAD}} \\
\midrule
\multirow{4}{*}{1} & DINOv2 & SubspaceAD & \textbf{+2.28}{\scriptsize$\pm$0.40} & \textbf{+2.16}{\scriptsize$\pm$0.58} & \textbf{+0.17}{\scriptsize$\pm$0.04} & \textbf{+4.70}{\scriptsize$\pm$0.70} & \textbf{+0.37}{\scriptsize$\pm$0.11} \\
 &  & DuoAD & \textbf{+2.71}{\scriptsize$\pm$0.59} & \textbf{+3.14}{\scriptsize$\pm$0.75} & \textbf{+0.48}{\scriptsize$\pm$0.03} & \textbf{-2.24}{\scriptsize$\pm$0.94} & \textbf{+1.52}{\scriptsize$\pm$0.11} \\
 & DINOv3 & SubspaceAD & \textbf{+4.57}{\scriptsize$\pm$0.53} & \textbf{+4.66}{\scriptsize$\pm$0.44} & \textbf{+0.39}{\scriptsize$\pm$0.06} & \textbf{+5.56}{\scriptsize$\pm$0.59} & \textbf{+2.10}{\scriptsize$\pm$0.21} \\
 &  & DuoAD & +0.02{\scriptsize$\pm$0.39} & +0.40{\scriptsize$\pm$0.38} & \textbf{+0.03}{\scriptsize$\pm$0.01} & \textbf{+5.73}{\scriptsize$\pm$0.47} & +0.01{\scriptsize$\pm$0.10} \\
\cmidrule(lr){1-8}
\multirow{4}{*}{2} & DINOv2 & SubspaceAD & \textbf{+1.79}{\scriptsize$\pm$0.32} & \textbf{+1.61}{\scriptsize$\pm$0.51} & \textbf{+0.19}{\scriptsize$\pm$0.03} & \textbf{+4.27}{\scriptsize$\pm$0.51} & \textbf{+0.32}{\scriptsize$\pm$0.06} \\
 &  & DuoAD & \textbf{+2.24}{\scriptsize$\pm$0.50} & \textbf{+2.77}{\scriptsize$\pm$0.70} & \textbf{+0.43}{\scriptsize$\pm$0.02} & \textbf{-2.46}{\scriptsize$\pm$0.75} & \textbf{+1.28}{\scriptsize$\pm$0.08} \\
 & DINOv3 & SubspaceAD & \textbf{+3.87}{\scriptsize$\pm$0.48} & \textbf{+3.79}{\scriptsize$\pm$0.41} & \textbf{+0.30}{\scriptsize$\pm$0.03} & \textbf{+4.44}{\scriptsize$\pm$0.60} & \textbf{+1.64}{\scriptsize$\pm$0.11} \\
 &  & DuoAD & +0.06{\scriptsize$\pm$0.31} & +0.39{\scriptsize$\pm$0.29} & +0.00{\scriptsize$\pm$0.02} & \textbf{+4.92}{\scriptsize$\pm$0.43} & +0.04{\scriptsize$\pm$0.08} \\
\cmidrule(lr){1-8}
\multirow{4}{*}{4} & DINOv2 & SubspaceAD & \textbf{+1.38}{\scriptsize$\pm$0.11} & \textbf{+1.31}{\scriptsize$\pm$0.24} & \textbf{+0.22}{\scriptsize$\pm$0.02} & \textbf{+3.21}{\scriptsize$\pm$0.39} & \textbf{+0.27}{\scriptsize$\pm$0.08} \\
 &  & DuoAD & \textbf{+1.86}{\scriptsize$\pm$0.24} & \textbf{+2.65}{\scriptsize$\pm$0.35} & \textbf{+0.43}{\scriptsize$\pm$0.02} & \textbf{-3.21}{\scriptsize$\pm$0.50} & \textbf{+1.15}{\scriptsize$\pm$0.10} \\
 & DINOv3 & SubspaceAD & \textbf{+3.30}{\scriptsize$\pm$0.30} & \textbf{+3.24}{\scriptsize$\pm$0.33} & \textbf{+0.29}{\scriptsize$\pm$0.02} & \textbf{+3.11}{\scriptsize$\pm$0.45} & \textbf{+1.40}{\scriptsize$\pm$0.08} \\
 &  & DuoAD & -0.19{\scriptsize$\pm$0.20} & +0.29{\scriptsize$\pm$0.20} & \textbf{+0.02}{\scriptsize$\pm$0.01} & \textbf{+3.50}{\scriptsize$\pm$0.45} & +0.04{\scriptsize$\pm$0.06} \\
\bottomrule
\end{tabular}
\end{table*}

\FloatBarrier
\subsection{VisA per Category Paired Differences}
\label{sec:supp_visa_categories}

\Cref{tab:visa_categories} breaks down the paired differences on VisA by category.

\begin{table*}[t]
\centering
\caption{LocalSubspaceAD minus SubspaceAD on VisA per category (DINOv2), in points, averaged over 1/2/4 shots: mean $\pm$ std over 10 seeds of the paired difference. Base: SubspaceAD's I-AUROC. Positive values show gain of our method. \textbf{Bold}: the same sign at every seed.}
\label{tab:visa_categories}
\begin{tabular}{lcccccc}
\toprule
Category & Base & $\Delta$ I-AUROC & $\Delta$ I-AUPR & $\Delta$ P-AUROC & $\Delta$ P-AUPR & $\Delta$ AUPRO \\
\midrule
\multicolumn{7}{l}{\textit{Multi-instance}} \\
\quad macaroni2 & 81.61 & +4.06{\scriptsize$\pm$3.41} & +6.57{\scriptsize$\pm$4.44} & +0.04{\scriptsize$\pm$0.05} & +4.33{\scriptsize$\pm$2.13} & -0.22{\scriptsize$\pm$0.47} \\
\quad macaroni1 & 93.92 & \textbf{+1.63}{\scriptsize$\pm$0.89} & \textbf{+2.11}{\scriptsize$\pm$1.09} & \textbf{+0.12}{\scriptsize$\pm$0.04} & +0.45{\scriptsize$\pm$1.44} & \textbf{+1.30}{\scriptsize$\pm$0.29} \\
\quad capsules & 96.85 & \textbf{+1.25}{\scriptsize$\pm$0.26} & \textbf{+0.71}{\scriptsize$\pm$0.17} & \textbf{+0.45}{\scriptsize$\pm$0.11} & \textbf{+19.06}{\scriptsize$\pm$3.94} & \textbf{+0.83}{\scriptsize$\pm$0.16} \\
\quad candle & 94.50 & \textbf{+0.98}{\scriptsize$\pm$0.36} & \textbf{+1.52}{\scriptsize$\pm$0.36} & \textbf{+0.08}{\scriptsize$\pm$0.03} & \textbf{+9.90}{\scriptsize$\pm$1.65} & -0.11{\scriptsize$\pm$0.10} \\
\cmidrule(lr){1-7}
\quad Group mean & 91.72 & \textbf{+1.98}{\scriptsize$\pm$0.76} & \textbf{+2.73}{\scriptsize$\pm$0.97} & \textbf{+0.17}{\scriptsize$\pm$0.02} & \textbf{+8.43}{\scriptsize$\pm$0.55} & \textbf{+0.45}{\scriptsize$\pm$0.15} \\
\midrule
\multicolumn{7}{l}{\textit{PCB}} \\
\quad pcb3 & 86.86 & \textbf{+3.99}{\scriptsize$\pm$1.69} & \textbf{+3.25}{\scriptsize$\pm$2.11} & \textbf{+1.70}{\scriptsize$\pm$0.17} & \textbf{+8.18}{\scriptsize$\pm$2.30} & \textbf{+2.14}{\scriptsize$\pm$0.60} \\
\quad pcb2 & 90.46 & \textbf{+3.62}{\scriptsize$\pm$1.62} & \textbf{+4.14}{\scriptsize$\pm$1.68} & +0.05{\scriptsize$\pm$0.10} & \textbf{+5.37}{\scriptsize$\pm$1.74} & \textbf{+0.73}{\scriptsize$\pm$0.19} \\
\quad pcb1 & 92.69 & \textbf{+2.42}{\scriptsize$\pm$0.96} & \textbf{+2.85}{\scriptsize$\pm$1.25} & \textbf{+0.06}{\scriptsize$\pm$0.06} & \textbf{+6.35}{\scriptsize$\pm$2.91} & +0.10{\scriptsize$\pm$0.46} \\
\quad pcb4 & 98.02 & \textbf{+1.11}{\scriptsize$\pm$0.58} & \textbf{+1.36}{\scriptsize$\pm$0.70} & \textbf{+0.67}{\scriptsize$\pm$0.20} & \textbf{+3.56}{\scriptsize$\pm$1.52} & +0.28{\scriptsize$\pm$0.51} \\
\cmidrule(lr){1-7}
\quad Group mean & 92.01 & \textbf{+2.78}{\scriptsize$\pm$0.89} & \textbf{+2.90}{\scriptsize$\pm$0.97} & \textbf{+0.62}{\scriptsize$\pm$0.06} & \textbf{+5.86}{\scriptsize$\pm$1.28} & \textbf{+0.81}{\scriptsize$\pm$0.26} \\
\midrule
\multicolumn{7}{l}{\textit{Single-instance}} \\
\quad chewinggum & 99.10 & \textbf{+0.55}{\scriptsize$\pm$0.07} & \textbf{+0.30}{\scriptsize$\pm$0.03} & \textbf{-0.06}{\scriptsize$\pm$0.02} & \textbf{-11.50}{\scriptsize$\pm$2.83} & \textbf{+0.43}{\scriptsize$\pm$0.16} \\
\quad pipe\_fryum & 95.79 & +0.44{\scriptsize$\pm$0.63} & +0.41{\scriptsize$\pm$0.43} & \textbf{+0.10}{\scriptsize$\pm$0.04} & \textbf{+4.20}{\scriptsize$\pm$0.89} & -0.01{\scriptsize$\pm$0.09} \\
\quad fryum & 97.10 & -1.11{\scriptsize$\pm$1.02} & -0.56{\scriptsize$\pm$0.52} & -0.17{\scriptsize$\pm$0.16} & -0.08{\scriptsize$\pm$0.82} & -0.16{\scriptsize$\pm$0.24} \\
\quad cashew & 98.42 & \textbf{-2.64}{\scriptsize$\pm$0.51} & \textbf{-1.39}{\scriptsize$\pm$0.28} & +0.01{\scriptsize$\pm$0.04} & \textbf{+4.44}{\scriptsize$\pm$1.36} & -0.09{\scriptsize$\pm$0.14} \\
\cmidrule(lr){1-7}
\quad Group mean & 97.60 & \textbf{-0.69}{\scriptsize$\pm$0.23} & \textbf{-0.31}{\scriptsize$\pm$0.13} & -0.03{\scriptsize$\pm$0.03} & -0.73{\scriptsize$\pm$0.82} & +0.04{\scriptsize$\pm$0.07} \\
\midrule
All VisA & 93.78 & \textbf{+1.36}{\scriptsize$\pm$0.39} & \textbf{+1.77}{\scriptsize$\pm$0.46} & \textbf{+0.25}{\scriptsize$\pm$0.03} & \textbf{+4.52}{\scriptsize$\pm$0.58} & \textbf{+0.44}{\scriptsize$\pm$0.08} \\
\bottomrule
\end{tabular}
\end{table*}

\pagebreak
\FloatBarrier
\section{Number of Components per Shot}
\label{sec:supp_kshot}

\begin{figure*}[t]
\centering
\includegraphics[width=0.6\linewidth]{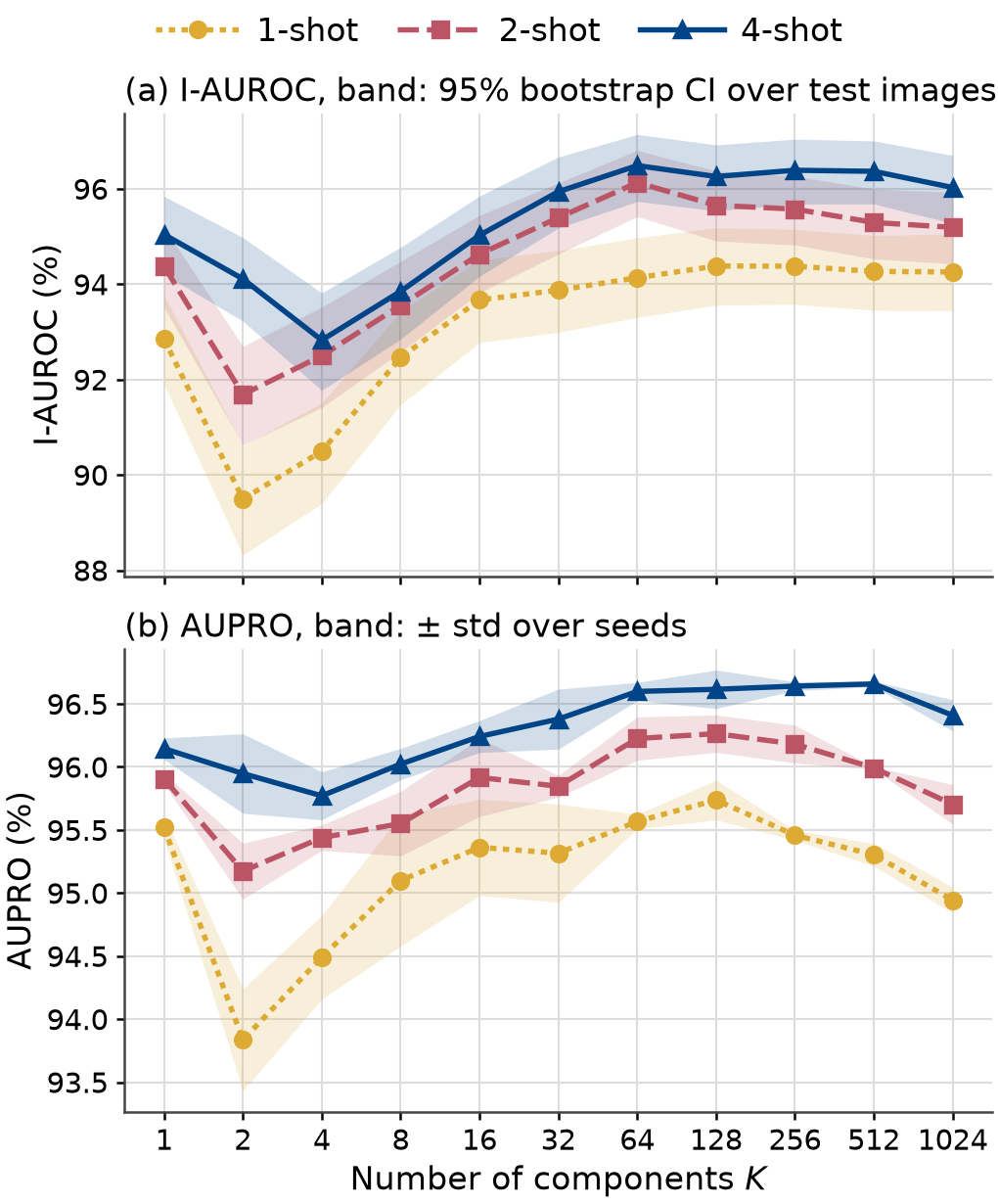}
\caption{Image AUROC and AUPRO of LocalSubspaceAD with different $K$ components by shots, VisA. Each point is the mean over the 12 categories and three seeds. I-AUROC bands are 95\,\% bootstrap confidence intervals over test images with 2,000 replicates. The AUPRO bands show $\pm$ one standard deviation across seeds. $K$ is capped at $\lfloor N/20 \rfloor$, where $N$ is the number of support patches (\cref{ssec:k-sweep}).}
\label{fig:k-sweep}
\end{figure*}

Per shot (\cref{fig:k-sweep}), the best $K$ for localization grows with the amount of support data: AUPRO peaks at $K=128$ for 1-shot ($95.74$, from $95.52$ at $K=1$) and 2-shot ($96.26$ over $95.91$), and at $K=512$ for 4-shot ($96.66$ over $96.15$), while at $K=128$ it improves over $K=1$ at every shot count ($96.15 \to 96.62$ at 4-shot). Image-level detection peaks at $K=128$ for 1-shot and at $K=64$ for 2- and 4-shot and stays roughly flat beyond, rising from $92.85$/$94.38$/$95.04$ at $K=1$ to $94.38$/$95.65$/$96.26$ at $K=128$ for 1/2/4 shots.

\pagebreak
\FloatBarrier
\section{Component Principal Directions}\label{sec:supp_dims}
\Cref{tab:q_k} shows the component principal directions, number of patches and fit times for SubspaceAD ($K=1$) and LocalSubspaceAD ($K=128$) using DINOv2. A median component keeps $\sim 310$ principal directions from $\sim 1000$ patches, drawn from multiple views of the same image (averaged across the three datasets and shots). $r_k$ rises with the number of shots. The median maximum number of retained directions $\max_k r_k$ is around two thirds of the retained directions with $K=1$. 

\begin{table*}[t]
\centering
\caption{Subspace sizes of LocalSubspaceAD (DINOv2-G, $D=1536$, $\tau=0.99$) on the three datasets: the $K=128$ local subspaces' numbers of directions $r_k$ against the single global subspace's $r$ ($K=1$), median over the categories and three seeds; \textit{Mean}: mean over 1/2/4 shots. $r_k$: median over the components, with the 10th and 90th percentiles (p10, p90) over the components in brackets. $n_k$: median cluster size, $N$ support patches. Fit: $K=1 \to K=128$ (k-means and the per-component eigendecompositions).}
\label{tab:q_k}
\begin{tabular}{llcccccc}
\toprule
Dataset & Shots & $r$ ($K=1$) & $r_k$ median [p10, p90] & $\max_k r_k$ & $n_k$ & $N$ & Fit (s)\\
\midrule
MVTec AD & 1 & 905 & 210 [98, 359] & 498 & 502 & 71k & 1.4 $\to$ 26 \\
 & 2 & 942 & 328 [168, 493] & 623 & 986 & 143k & 3.2 $\to$ 31 \\
 & 4 & 978 & 460 [269, 627] & 741 & 1984 & 286k & 6.5 $\to$ 43 \\
\cmidrule(l){2-8}
 & \textit{Mean} & 942 & 332 [178, 493] & 621 &  &  & 3.7 $\to$ 34 \\
\midrule
VisA & 1 & 952 & 184 [98, 366] & 516 & 420 & 71k & 1.3 $\to$ 16 \\
 & 2 & 984 & 290 [161, 504] & 646 & 836 & 143k & 2.5 $\to$ 22 \\
 & 4 & 1004 & 437 [245, 644] & 755 & 1676 & 286k & 4.8 $\to$ 33 \\
\cmidrule(l){2-8}
 & \textit{Mean} & 980 & 304 [168, 504] & 639 &  &  & 2.9 $\to$ 24 \\
\midrule
Real-IAD & 1 & 886 & 180 [98, 294] & 420 & 445 & 71k & 1.1 $\to$ 15 \\
 & 2 & 927 & 283 [154, 430] & 560 & 877 & 143k & 2.2 $\to$ 20 \\
 & 4 & 952 & 407 [235, 570] & 694 & 1787 & 286k & 4.2 $\to$ 31 \\
\cmidrule(l){2-8}
 & \textit{Mean} & 921 & 290 [162, 431] & 558 &  &  & 2.5 $\to$ 22 \\
\bottomrule
\end{tabular}
\end{table*}

\FloatBarrier
\pagebreak
\section{Explained-Variance Threshold} \label{sec:supp_tau}
\Cref{tab:tau} shows the effects of the explained-variance threshold $\tau$, averaged over shots and per shot count. From $\tau=0.95$ to $0.995$ all differences stay within $0.32$ points, and the per-shot I-AUROC and AUPRO differences within $0.37$; $\tau=0.90$ loses up to $0.83$ I-AUROC and $0.51$ AUPRO (1-shot). Per shot, P-AUPR is noisier and depends on the support set size: a lower $\tau$ costs up to $1.89$ points at 1-shot but gains up to $1.24$ at 4-shot, which largely cancels in the average. The drop at $\tau=0.999$ does not shrink consistently with more support data: $-1.33$ I-AUROC and $-0.29$ AUPRO at 1-shot, $-1.96$ and $-0.55$ at 2-shot, $-1.02$ and $-0.78$ at 4-shot.

\begin{table}[t]
\centering
\caption{Effect of the explained-variance threshold $\tau$ on VisA (DINOv2, $K=128$), three seeds, averaged over 1/2/4 shots and per shot count. Default ($\tau=0.99$, bold): absolute values; other rows: paired difference to the default in points. Mean $\pm$ std over seeds.}
\label{tab:tau}
\setlength{\tabcolsep}{5pt}
\begin{tabular}{l cccc}
\toprule
$\tau$ & Average & 1-shot & 2-shot & 4-shot \\
\midrule
\multicolumn{5}{l}{\textit{I-AUROC}} \\
0.90 & $-$0.67{\scriptsize$\pm$0.23} & $-$0.83{\scriptsize$\pm$0.63} & $-$0.71{\scriptsize$\pm$0.03} & $-$0.47{\scriptsize$\pm$0.10} \\
0.95 & $-$0.32{\scriptsize$\pm$0.10} & $-$0.32{\scriptsize$\pm$0.29} & $-$0.28{\scriptsize$\pm$0.02} & $-$0.37{\scriptsize$\pm$0.13} \\
0.98 & $-$0.06{\scriptsize$\pm$0.02} & $-$0.05{\scriptsize$\pm$0.07} & $-$0.02{\scriptsize$\pm$0.16} & $-$0.10{\scriptsize$\pm$0.09} \\
\textbf{0.99} & \textbf{95.43}{\scriptsize$\pm$0.21} & \textbf{94.38}{\scriptsize$\pm$0.12} & \textbf{95.65}{\scriptsize$\pm$0.53} & \textbf{96.26}{\scriptsize$\pm$0.20} \\
0.995 & +0.01{\scriptsize$\pm$0.27} & $-$0.07{\scriptsize$\pm$0.39} & $-$0.08{\scriptsize$\pm$0.20} & +0.19{\scriptsize$\pm$0.27} \\
0.999 & $-$1.44{\scriptsize$\pm$0.31} & $-$1.33{\scriptsize$\pm$0.60} & $-$1.96{\scriptsize$\pm$0.25} & $-$1.02{\scriptsize$\pm$0.18} \\
\midrule
\multicolumn{5}{l}{\textit{P-AUPR}} \\
0.90 & $-$0.40{\scriptsize$\pm$0.51} & $-$1.89{\scriptsize$\pm$0.44} & $-$0.22{\scriptsize$\pm$0.62} & +0.90{\scriptsize$\pm$0.52} \\
0.95 & +0.04{\scriptsize$\pm$0.36} & $-$1.32{\scriptsize$\pm$0.87} & +0.22{\scriptsize$\pm$0.15} & +1.24{\scriptsize$\pm$0.14} \\
0.98 & +0.05{\scriptsize$\pm$0.49} & $-$0.78{\scriptsize$\pm$0.45} & +0.45{\scriptsize$\pm$0.66} & +0.49{\scriptsize$\pm$0.42} \\
\textbf{0.99} & \textbf{43.52}{\scriptsize$\pm$0.39} & \textbf{42.29}{\scriptsize$\pm$0.45} & \textbf{43.76}{\scriptsize$\pm$0.33} & \textbf{44.51}{\scriptsize$\pm$0.44} \\
0.995 & +0.20{\scriptsize$\pm$0.66} & +0.31{\scriptsize$\pm$0.46} & +0.49{\scriptsize$\pm$1.24} & $-$0.19{\scriptsize$\pm$0.71} \\
0.999 & $-$1.97{\scriptsize$\pm$0.39} & $-$1.53{\scriptsize$\pm$0.19} & $-$1.26{\scriptsize$\pm$0.90} & $-$3.11{\scriptsize$\pm$2.08} \\
\midrule
\multicolumn{5}{l}{\textit{AUPRO}} \\
0.90 & $-$0.32{\scriptsize$\pm$0.04} & $-$0.51{\scriptsize$\pm$0.07} & $-$0.29{\scriptsize$\pm$0.05} & $-$0.16{\scriptsize$\pm$0.15} \\
0.95 & $-$0.14{\scriptsize$\pm$0.03} & $-$0.32{\scriptsize$\pm$0.11} & $-$0.10{\scriptsize$\pm$0.15} & $-$0.01{\scriptsize$\pm$0.13} \\
0.98 & $-$0.02{\scriptsize$\pm$0.05} & $-$0.15{\scriptsize$\pm$0.11} & +0.05{\scriptsize$\pm$0.10} & +0.05{\scriptsize$\pm$0.17} \\
\textbf{0.99} & \textbf{96.21}{\scriptsize$\pm$0.08} & \textbf{95.74}{\scriptsize$\pm$0.16} & \textbf{96.26}{\scriptsize$\pm$0.15} & \textbf{96.62}{\scriptsize$\pm$0.15} \\
0.995 & $-$0.02{\scriptsize$\pm$0.23} & +0.06{\scriptsize$\pm$0.15} & $-$0.11{\scriptsize$\pm$0.31} & $-$0.02{\scriptsize$\pm$0.28} \\
0.999 & $-$0.54{\scriptsize$\pm$0.03} & $-$0.29{\scriptsize$\pm$0.15} & $-$0.55{\scriptsize$\pm$0.13} & $-$0.78{\scriptsize$\pm$0.22} \\
\bottomrule
\end{tabular}
\end{table}

\FloatBarrier
\section{Component Selection} \label{sec:supp_selection}
\begin{figure}[t]
\centering
\includegraphics[width=0.6\linewidth]{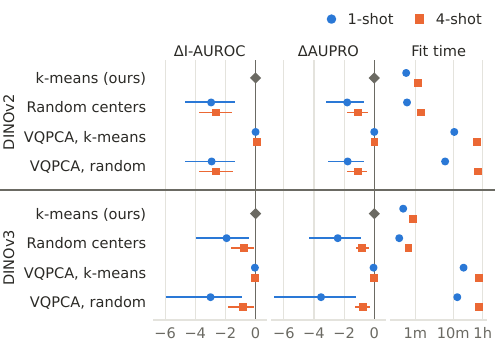}
\caption{Paired difference to k-means in image AUROC and AUPRO (with 95\,\% CI) and median fit time (in log scale) of LocalSubspaceAD ($K=128$) with different component selection methods on DINOv2 and DINOv3, VisA, three seeds. VQPCA: reassigning patches to the component with the smallest reprojection residual ($d_k$) and refitting from random initial centers (random) and from k-means. Absolute values are in \cref{tab:initialization}.}
\label{fig:initialization}
\end{figure}

\Cref{fig:initialization} compares k-means with other ways of selecting the components; the numerical values are in \cref{tab:initialization}. The experiments show that k-means-based component selection is sufficient, with no other method meaningfully improving on k-means-based component selection. We use randomly selected cluster centers ("Random centers") to show that semantic clustering is necessary, whereas iteration-based VQPCA \citeS{kambhatlaDimensionReductionLocal1997} does not improve performance and is much slower. The values of the fit times were averaged across all seeds, shots and datasets in the experiment. The fit times are shown in log scale, showing the order of magnitude differences between the k-means and iteration-based assignment methods. The exact fit times are displayed in \cref{tab:initialization} 

\begin{table}[t]
\centering
\caption{Cluster initialization of LocalSubspaceAD ($K=128$) on VisA, mean $\pm$ std over 3 seeds. Bold: best per backbone and column. Fit: estimated times, median fit time per run, on a shared workload.}
\label{tab:initialization}
\begin{tabular}{llcccc}
\toprule
Backbone & Method & I-AUROC & P-AUPR & AUPRO & Fit \\
\midrule
\multicolumn{6}{l}{\textit{1-shot}} \\
\midrule
DINOv2 & k-means (ours) & \textbf{94.38}{\scriptsize$\pm$0.12} & \textbf{42.29}{\scriptsize$\pm$0.45} & \textbf{95.74}{\scriptsize$\pm$0.16} & 34\,s \\
 & Random centers, no iteration & 91.43{\scriptsize$\pm$1.44} & 33.75{\scriptsize$\pm$2.25} & 93.94{\scriptsize$\pm$0.85} & 36\,s \\
 & VQPCA, k-means start & \textbf{94.38}{\scriptsize$\pm$0.12} & \textbf{42.29}{\scriptsize$\pm$0.45} & \textbf{95.74}{\scriptsize$\pm$0.16} & 11\,min \\
 & VQPCA, random start & 91.46{\scriptsize$\pm$1.37} & 33.74{\scriptsize$\pm$2.18} & 93.97{\scriptsize$\pm$0.82} & 6\,min \\
\midrule
DINOv3 & k-means (ours) & \textbf{93.68}{\scriptsize$\pm$0.68} & \textbf{55.04}{\scriptsize$\pm$0.70} & \textbf{94.74}{\scriptsize$\pm$0.20} & 28\,s \\
 & Random centers, no iteration & 91.75{\scriptsize$\pm$1.08} & 47.83{\scriptsize$\pm$1.17} & 92.31{\scriptsize$\pm$0.56} & 22\,s \\
 & VQPCA, k-means start & 93.63{\scriptsize$\pm$0.64} & 54.66{\scriptsize$\pm$1.08} & 94.69{\scriptsize$\pm$0.25} & 19\,min \\
 & VQPCA, random start & 90.69{\scriptsize$\pm$1.10} & 47.04{\scriptsize$\pm$1.91} & 91.21{\scriptsize$\pm$0.44} & 13\,min \\
\midrule
\multicolumn{6}{l}{\textit{4-shot}} \\
\midrule
DINOv2 & k-means (ours) & 96.26{\scriptsize$\pm$0.20} & 44.51{\scriptsize$\pm$0.44} & 96.62{\scriptsize$\pm$0.15} & 69\,s \\
 & Random centers, no iteration & 93.60{\scriptsize$\pm$1.07} & 38.49{\scriptsize$\pm$2.36} & 95.54{\scriptsize$\pm$0.36} & 83\,s \\
 & VQPCA, k-means start & \textbf{96.34}{\scriptsize$\pm$0.26} & \textbf{45.01}{\scriptsize$\pm$0.77} & \textbf{96.63}{\scriptsize$\pm$0.15} & 43\,min \\
 & VQPCA, random start & 93.64{\scriptsize$\pm$0.77} & 38.28{\scriptsize$\pm$1.61} & 95.53{\scriptsize$\pm$0.29} & 45\,min \\
\midrule
DINOv3 & k-means (ours) & \textbf{96.15}{\scriptsize$\pm$0.20} & \textbf{54.81}{\scriptsize$\pm$0.53} & \textbf{95.76}{\scriptsize$\pm$0.01} & 53\,s \\
 & Random centers, no iteration & 95.38{\scriptsize$\pm$0.22} & 51.69{\scriptsize$\pm$0.78} & 94.97{\scriptsize$\pm$0.13} & 39\,s \\
 & VQPCA, k-means start & 96.14{\scriptsize$\pm$0.19} & 54.75{\scriptsize$\pm$0.46} & 95.72{\scriptsize$\pm$0.03} & 47\,min \\
 & VQPCA, random start & 95.31{\scriptsize$\pm$0.35} & 51.67{\scriptsize$\pm$0.53} & 94.99{\scriptsize$\pm$0.06} & 47\,min \\
\bottomrule
\end{tabular}
\end{table}

\FloatBarrier
\section{MPPCA -- A Probabilistic Extension of LocalSubspaceAD}
\label{sec:probabilistic}
\label{sec:supp_ppca}

In this section, we develop an extension of LocalSubspaceAD based on a mixture of probabilistic PCA (MPPCA). We find that scoring with the reprojection residual is comparable to scoring with a full probabilistic density model. 

Turning our method into a Gaussian density model would entail adding three components: (i) a penalty for deviations inside the subspace, which the residual ignores by construction, (ii) a correction that makes components with different spread and dimension comparable, and (iii) an assignment of patches to components by likelihood instead of by Euclidean distance to the cluster centers. None of the three improves detection consistently, and estimating the noise variance per component, as a standard MPPCA does, makes detection collapse.

\subsection{Model and score}
We model each component $k$ as a PPCA~\citeS{tippingProbabilisticPrincipalComponent1999,tippingMixturesProbabilisticPrincipal1999} with the mean $\mu_k$, basis $B_k$ and eigenvalues $\Lambda_k = \text{diag}(\lambda_{k,1}, \dots, \lambda_{k,r_k})$ of the local PCA (\cref{eq:local_pca,eq:tau}) and the pooled noise variance $\sigma^2$ of \cref{eq:sigma_pooled}:
\begin{equation} \label{eq:probabilistic_model}
    x_p=\mu_k + B_k z_p+\epsilon_p, \quad
    z_p \sim \mathcal{N}(0, \Lambda_k - \sigma^2 I_{r_k}), \quad
    \epsilon_p \sim \mathcal{N}(0, \sigma^2I_D).
\end{equation}
Here $z_p$ are the latent coordinates in the subspace and $\epsilon_p$ is isotropic noise. Integrating out $z_p$ gives the density $p_k(x_p) = \mathcal{N}(x_p; \mu_k, C_k)$ with $C_k = B_k \Lambda_k B_k^\mathsf{T} + \sigma^2 (I_D - B_k B_k^\mathsf{T})$, which keeps the sample variance along the $r_k$ retained directions and assigns $\sigma^2$ to each of the other $D-r_k$. For a fixed $r_k$, $B_k$ is also the maximum likelihood (ML) subspace of the PPCA~\citeS{tippingProbabilisticPrincipalComponent1999}, and $\sigma^2$ is the ML noise variance shared by all components, except that it averages only over the $R_k - r_k$ discarded directions that carry sample variance, where $R_k = \min(|X_k| - 1, D)$ is the maximal rank of $\Sigma_k$. Because $\sigma^2$ is pooled, a retained eigenvalue can fall below it, which would make the latent variance $\lambda_{k,j} - \sigma^2$ negative. We therefore floor the retained eigenvalues at $\max(\lambda_{k,j}, \sigma^2)$, the ML solution given $\sigma^2$, and use the floored values in $\Lambda_k$ from here on.

Each patch is scored by its best component, $\min_k \text{NLL}_k(x_p)$, where $\text{NLL}_k(x_p) = -2\log p_k(x_p)$ is the negative log-likelihood under component $k$ (scaled by $2$, which does not change the ranking of patches):
\begin{equation} \label{eq:nll_detailed}
  \text{NLL}_k(x_p) = m_k(x_p) + d_k(x_p) / \sigma^2 + c_k + D\log 2\pi.
\end{equation}
In this expression, $d_k/\sigma^2$ is the reprojection residual (\cref{eq:scoring}) divided by the noise variance, 
\begin{equation} \label{eq:mahal}
m_k(x_p) = \|\Lambda_k^{-1/2} B_k^\mathsf{T} (x_p-\mu_k)\|^2
\end{equation} 
is the squared Mahalanobis distance inside the subspace (i), and 
\begin{equation} \label{eq:logdet}
c_k = \log\det C_k = \sum_{j=1}^{r_k}\log\lambda_{k,j} + (D-r_k)\log\sigma^2
\end{equation}
is the log-determinant (ii), which is larger for components with more spread or more retained dimensions. Because $\sigma^2$ is the same for all components, $d_k/\sigma^2$ alone ranks patches exactly as the residual score does, so any difference between the two scores comes from $m_k$ and $c_k$. 
\newpage
\subsection{Density terms i) and ii)}
\Cref{tab:ppca_decomposition} compares adding the terms one at a time to the residual score on the same fit. Choosing the component by the NLL, adding $m_k$ and adding $c_k$ each change every metric by at most $0.12$ points. The full NLL is $0.10$ AUPRO above the residual score on MVTec AD, $0.04$ AUPRO below it on Real-IAD and $0.01$ to $0.10$ points below it in image-level detection on all datasets.

\begin{table*}[t]
\centering
\caption{MPPCA compared to LocalSubspaceAD ($K=128$), averaged over 1/2/4 shots. First row per backbone: ours, mean $\pm$ std over 10 seeds; other rows: paired difference to ours, in points. \textbf{Bold}: the same sign at every seed and shot count. DINOv2: each row adds one term of the NLL \eqref{eq:nll_detailed} to the row above; the last row is the full mixture likelihood $-\log\sum_k \pi_k p_k$. DINOv3: ways of combining the layers' residuals; MPPCA per layer is fitted separately.}
\label{tab:ppca_decomposition}
\setlength{\tabcolsep}{2.5pt}
\footnotesize
\begin{adjustbox}{max width=\linewidth}
\begin{tabular}{ll@{\hspace{5pt}}ccc@{\hspace{5pt}}ccc@{\hspace{5pt}}ccc}
\toprule
& & \multicolumn{3}{c}{MVTec AD} & \multicolumn{3}{c}{VisA} & \multicolumn{3}{c}{Real-IAD} \\
\cmidrule(lr){3-5} \cmidrule(lr){6-8} \cmidrule(lr){9-11}
Backbone & Score & I-AUROC & P-AUPR & AUPRO & I-AUROC & P-AUPR & AUPRO & I-AUROC & P-AUPR & AUPRO \\
\midrule
\multirow{5}{*}{DINOv2} & LocalSubspaceAD, $\min_k d_k$ & 97.97{\tiny$\pm$0.46} & 60.19{\tiny$\pm$0.68} & 95.26{\tiny$\pm$0.18} & 95.13{\tiny$\pm$0.41} & 43.52{\tiny$\pm$0.39} & 96.15{\tiny$\pm$0.07} & 92.00{\tiny$\pm$0.24} & 49.77{\tiny$\pm$0.42} & 98.00{\tiny$\pm$0.07} \\
\cmidrule(lr){2-11}
 & $d_k$ at the NLL's component & -0.00{\tiny$\pm$0.00} & \textbf{-0.09}{\tiny$\pm$0.02} & \textbf{-0.01}{\tiny$\pm$0.00} & -0.02{\tiny$\pm$0.01} & \textbf{-0.12}{\tiny$\pm$0.03} & -0.01{\tiny$\pm$0.00} & -0.01{\tiny$\pm$0.00} & -0.06{\tiny$\pm$0.02} & \textbf{-0.01}{\tiny$\pm$0.00} \\
 & $+\,m_k$ ($\min_k d_k/\sigma^2 + m_k$) & -0.01{\tiny$\pm$0.02} & -0.11{\tiny$\pm$0.06} & +0.04{\tiny$\pm$0.01} & -0.09{\tiny$\pm$0.03} & -0.03{\tiny$\pm$0.07} & -0.01{\tiny$\pm$0.01} & \textbf{-0.10}{\tiny$\pm$0.03} & -0.03{\tiny$\pm$0.04} & \textbf{-0.04}{\tiny$\pm$0.01} \\
 & $+\,c_k$: MPPCA, $\min_k \mathrm{NLL}_k$ & -0.01{\tiny$\pm$0.02} & -0.04{\tiny$\pm$0.06} & \textbf{+0.10}{\tiny$\pm$0.02} & -0.08{\tiny$\pm$0.03} & +0.05{\tiny$\pm$0.08} & +0.00{\tiny$\pm$0.01} & \textbf{-0.10}{\tiny$\pm$0.03} & +0.03{\tiny$\pm$0.04} & \textbf{-0.04}{\tiny$\pm$0.01} \\
 & Mixture, $-\log\sum_k \pi_k p_k$ & -0.01{\tiny$\pm$0.02} & -0.04{\tiny$\pm$0.06} & \textbf{+0.09}{\tiny$\pm$0.02} & -0.08{\tiny$\pm$0.03} & +0.05{\tiny$\pm$0.08} & +0.00{\tiny$\pm$0.01} & \textbf{-0.10}{\tiny$\pm$0.03} & +0.03{\tiny$\pm$0.04} & \textbf{-0.04}{\tiny$\pm$0.01} \\
\midrule
\multirow{4}{*}{DINOv3} & LocalSubspaceAD, $\min_k \sum_l d_{l,k}/\sigma^2_l$ & 97.59{\tiny$\pm$0.76} & 71.23{\tiny$\pm$0.73} & 95.74{\tiny$\pm$0.22} & 94.76{\tiny$\pm$0.46} & 55.19{\tiny$\pm$0.34} & 95.33{\tiny$\pm$0.07} & 89.70{\tiny$\pm$0.32} & 57.13{\tiny$\pm$0.42} & 96.71{\tiny$\pm$0.09} \\
\cmidrule(lr){2-11}
 & Raw sum, $\min_k \sum_l d_{l,k}$ & \textbf{-0.65}{\tiny$\pm$0.23} & \textbf{-0.84}{\tiny$\pm$0.42} & \textbf{-0.96}{\tiny$\pm$0.07} & \textbf{-2.81}{\tiny$\pm$0.28} & +0.00{\tiny$\pm$0.37} & \textbf{-2.06}{\tiny$\pm$0.16} & \textbf{-2.59}{\tiny$\pm$0.21} & \textbf{-2.61}{\tiny$\pm$0.25} & \textbf{-1.51}{\tiny$\pm$0.08} \\
 & Standardized per layer (zsum) & -0.09{\tiny$\pm$0.05} & +0.03{\tiny$\pm$0.10} & \textbf{-0.21}{\tiny$\pm$0.02} & \textbf{-0.36}{\tiny$\pm$0.08} & \textbf{+0.38}{\tiny$\pm$0.08} & \textbf{-0.16}{\tiny$\pm$0.02} & \textbf{-0.42}{\tiny$\pm$0.06} & -0.12{\tiny$\pm$0.08} & \textbf{-0.16}{\tiny$\pm$0.01} \\
 & MPPCA per layer, $\min_k \sum_l \mathrm{NLL}_{l,k}$ & -0.00{\tiny$\pm$0.04} & -0.00{\tiny$\pm$0.03} & \textbf{+0.09}{\tiny$\pm$0.01} & +0.02{\tiny$\pm$0.02} & -0.10{\tiny$\pm$0.03} & +0.01{\tiny$\pm$0.02} & +0.07{\tiny$\pm$0.03} & -0.03{\tiny$\pm$0.04} & +0.05{\tiny$\pm$0.01} \\
\bottomrule
\end{tabular}
\end{adjustbox}
\end{table*}

On DINOv3 (lower part of \cref{tab:ppca_decomposition}, \textit{MPPCA per layer}), the NLL per layer also changes the metrics by at most $0.10$ points. What matters for this backbone is that the residuals of the layers are made comparable before they are summed (\cref{eq:scoring_layerwise}): summing the raw residuals instead of dividing them by the pooled noise variance $\sigma_l^2$ of each layer $l$ costs $0.65$ to $2.81$ I-AUROC and $0.96$ to $2.06$ AUPRO, and standardizing them per layer using mean and standard deviation computed from the training data costs $0.09$ to $0.42$ I-AUROC.

\FloatBarrier
\subsection{Number of MPPCA components}

We investigate the performance of MPPCA based on varying numbers of components $K$ in \cref{tab:ppca_ksweep}. LocalSubspaceAD reaches comparable performance as MPPCA, losing at most $0.15$ points across all investigated $K$. The gap to LocalSubspaceAD is most pronounced with $4\leq K \leq 32$ for I-AUROC and $1\leq K \leq 16$ for P-AUPR.

\begin{table*}[t]
\centering
\caption{MPPCA NLL minus the LocalSubspaceAD residual score on the same fit per number of components $K$, VisA, averaged over 1/2/4 shots: mean $\pm$ std over 3 seeds of the paired difference. Left: absolute values of LocalSubspaceAD. \textbf{Bold}: the same sign at every seed and shot count.}
\label{tab:ppca_ksweep}
\begin{tabular}{r@{\hspace{5pt}}ccc@{\hspace{5pt}}ccc}
\toprule
 & \multicolumn{3}{c}{LocalSubspaceAD} & \multicolumn{3}{c}{MPPCA minus LocalSubspaceAD} \\
\cmidrule(lr){2-4} \cmidrule(lr){5-7}
$K$ & I-AUROC & P-AUPR & AUPRO & I-AUROC & P-AUPR & AUPRO \\
\midrule
1 & 94.09{\scriptsize$\pm$0.42} & 39.12{\scriptsize$\pm$0.16} & 95.86{\scriptsize$\pm$0.06} & -0.04{\scriptsize$\pm$0.02} & \textbf{-1.08}{\scriptsize$\pm$0.16} & \textbf{-0.46}{\scriptsize$\pm$0.02} \\
2 & 91.76{\scriptsize$\pm$0.89} & 35.67{\scriptsize$\pm$0.82} & 94.99{\scriptsize$\pm$0.10} & +0.15{\scriptsize$\pm$0.07} & \textbf{-0.73}{\scriptsize$\pm$0.16} & +0.09{\scriptsize$\pm$0.05} \\
4 & 91.94{\scriptsize$\pm$0.70} & 35.96{\scriptsize$\pm$0.62} & 95.23{\scriptsize$\pm$0.16} & \textbf{-0.43}{\scriptsize$\pm$0.10} & \textbf{-1.45}{\scriptsize$\pm$0.16} & +0.06{\scriptsize$\pm$0.06} \\
8 & 93.29{\scriptsize$\pm$0.60} & 37.68{\scriptsize$\pm$0.61} & 95.56{\scriptsize$\pm$0.25} & \textbf{-0.60}{\scriptsize$\pm$0.12} & \textbf{-1.17}{\scriptsize$\pm$0.29} & +0.07{\scriptsize$\pm$0.09} \\
16 & 94.44{\scriptsize$\pm$0.24} & 39.61{\scriptsize$\pm$0.34} & 95.84{\scriptsize$\pm$0.04} & \textbf{-0.58}{\scriptsize$\pm$0.15} & \textbf{-0.71}{\scriptsize$\pm$0.18} & +0.09{\scriptsize$\pm$0.01} \\
32 & 95.07{\scriptsize$\pm$0.21} & 41.74{\scriptsize$\pm$0.46} & 95.85{\scriptsize$\pm$0.22} & \textbf{-0.24}{\scriptsize$\pm$0.12} & -0.21{\scriptsize$\pm$0.13} & \textbf{+0.16}{\scriptsize$\pm$0.06} \\
64 & 95.58{\scriptsize$\pm$0.22} & 43.03{\scriptsize$\pm$0.44} & 96.13{\scriptsize$\pm$0.09} & -0.10{\scriptsize$\pm$0.03} & +0.02{\scriptsize$\pm$0.07} & \textbf{+0.06}{\scriptsize$\pm$0.01} \\
128 & 95.43{\scriptsize$\pm$0.21} & 43.52{\scriptsize$\pm$0.39} & 96.21{\scriptsize$\pm$0.08} & \textbf{-0.07}{\scriptsize$\pm$0.02} & +0.06{\scriptsize$\pm$0.09} & +0.00{\scriptsize$\pm$0.01} \\
256 & 95.45{\scriptsize$\pm$0.26} & 44.62{\scriptsize$\pm$0.08} & 96.09{\scriptsize$\pm$0.06} & -0.05{\scriptsize$\pm$0.03} & -0.03{\scriptsize$\pm$0.08} & -0.02{\scriptsize$\pm$0.01} \\
512 & 95.31{\scriptsize$\pm$0.28} & 45.31{\scriptsize$\pm$0.43} & 95.98{\scriptsize$\pm$0.03} & -0.07{\scriptsize$\pm$0.04} & -0.05{\scriptsize$\pm$0.11} & \textbf{-0.05}{\scriptsize$\pm$0.01} \\
1024 & 95.16{\scriptsize$\pm$0.27} & 44.94{\scriptsize$\pm$0.25} & 95.69{\scriptsize$\pm$0.08} & -0.04{\scriptsize$\pm$0.03} & \textbf{-0.12}{\scriptsize$\pm$0.03} & \textbf{-0.04}{\scriptsize$\pm$0.01} \\
\bottomrule
\end{tabular}
\end{table*}

\subsection{Noise variance per component}
In parallel to LocalSubspaceAD's noise-based weighting in the layerwise DINOv3 variant, pooling the $\sigma^2$ is essential (\cref{tab:ppca_noise}) for performance in MPPCA. Using a local noise variance $\sigma^2_k$ per component, estimated from the residual variance of only that component (as in a standard MPPCA), detection collapses at 1-shot: the MPPCA loses up to $22$ I-AUROC and $26.5$ AUPRO points, and the residual weighted by $\sigma^2_k$ leads to decreasing performance of up to $30$ I-AUROC and $81$ AUPRO points (DINOv3).

\begin{table}[t]
\centering
\caption{One pooled noise variance against one per component on VisA ($K=128$, same fit). First row per backbone: full MPPCA, mean $\pm$ std over 3 seeds; other rows: paired difference to LocalSubspaceAD, in points. \textbf{Bold}: the same sign at every seed.}
\label{tab:ppca_noise}
\setlength{\tabcolsep}{2.5pt}
\footnotesize
\begin{adjustbox}{max width=\linewidth}
\begin{tabular}{l@{\hspace{5pt}}cc@{\hspace{5pt}}cc}
\toprule
& \multicolumn{2}{c}{1-shot} & \multicolumn{2}{c}{4-shot} \\
\cmidrule(lr){2-3} \cmidrule(lr){4-5}
Score & I-AUROC & AUPRO & I-AUROC & AUPRO \\
\midrule
\multicolumn{5}{l}{\textit{DINOv2}} \\
LocalSubspaceAD, $\min_k d_k$ & 94.38{\tiny$\pm$0.12} & 95.74{\tiny$\pm$0.16} & 96.26{\tiny$\pm$0.20} & 96.62{\tiny$\pm$0.15} \\
\cmidrule(lr){1-5}
MPPCA, pooled $\sigma^2$ & \textbf{-0.10}{\tiny$\pm$0.05} & \textbf{-0.04}{\tiny$\pm$0.01} & \textbf{-0.08}{\tiny$\pm$0.02} & +0.04{\tiny$\pm$0.03} \\
$\min_k d_k/\sigma^2_k$, own $\sigma^2_k$ & \textbf{-26.05}{\tiny$\pm$3.95} & \textbf{-53.89}{\tiny$\pm$4.35} & \textbf{-7.35}{\tiny$\pm$2.25} & \textbf{-4.77}{\tiny$\pm$1.86} \\
MPPCA, own $\sigma^2_k$ & \textbf{-22.25}{\tiny$\pm$3.93} & \textbf{-26.58}{\tiny$\pm$1.89} & \textbf{-7.04}{\tiny$\pm$1.93} & \textbf{-2.89}{\tiny$\pm$0.79} \\
\midrule
\multicolumn{5}{l}{\textit{DINOv3}} \\
$\min_k \sum_l d_{l,k}/\sigma^2_l$  & 93.68{\tiny$\pm$0.68} & 94.74{\tiny$\pm$0.20} & 96.15{\tiny$\pm$0.20} & 95.76{\tiny$\pm$0.01} \\
\cmidrule(lr){1-5}
$+\,m_k$ & \textbf{-0.04}{\tiny$\pm$0.02} & \textbf{-0.07}{\tiny$\pm$0.02} & +0.02{\tiny$\pm$0.05} & \textbf{+0.04}{\tiny$\pm$0.01} \\
MPPCA per layer, pooled $\sigma^2_l$ & -0.03{\tiny$\pm$0.05} & \textbf{-0.07}{\tiny$\pm$0.02} & +0.06{\tiny$\pm$0.07} & \textbf{+0.07}{\tiny$\pm$0.01} \\
$\min_k \sum_l d_{l,k}/\sigma^2_{l,k}$, own $\sigma^2_{l,k}$ & \textbf{-30.23}{\tiny$\pm$4.02} & \textbf{-81.13}{\tiny$\pm$2.99} & \textbf{-16.78}{\tiny$\pm$2.02} & \textbf{-38.14}{\tiny$\pm$6.77} \\
MPPCA per layer, own $\sigma^2_{l,k}$ & \textbf{-14.59}{\tiny$\pm$2.04} & \textbf{-23.11}{\tiny$\pm$0.24} & \textbf{-10.93}{\tiny$\pm$0.83} & \textbf{-7.96}{\tiny$\pm$0.59} \\
\bottomrule
\end{tabular}
\end{adjustbox}
\end{table}

\subsection{Likelihood-based component assignment (iii)}
\label{ssec:likelihood_assignment}
Performing PCA on the patches in each cluster does not give the maximum likelihood (ML) estimate of the mixture. We use expectation maximization (EM)~\citeS{tippingMixturesProbabilisticPrincipal1999} to estimate the distribution parameters instead. Using EM initialized from the k-means clusters and scoring by the NLL ($K=128$, VisA, \cref{tab:initialization_mppca_em}) performs worse as compared to the k-means+PCA-based variant of MPPCA on every metric: in particular by $2.48$ I-AUROC at 1-shot and $1.30$ at 4-shot on DINOv2 ($0.71$ and $0.63$ on DINOv3). 

\begin{table*}[t]
\centering
\caption{Cluster initialization of MPPCA (NLL score, $K=128$) on VisA: k-means+PCA against EM started from k-means, mean $\pm$ std over 3 seeds. Bold: best per backbone and column. Fit: median fit time per run; the k-means fits were timed in separate runs.}
\label{tab:initialization_mppca_em}
\begin{adjustbox}{max width=\linewidth}
\begin{tabular}{llcccccccc}
\toprule
Backbone & Method & \multicolumn{4}{c}{1-shot} & \multicolumn{4}{c}{4-shot} \\
\cmidrule(lr){3-6} \cmidrule(lr){7-10}
 & & I-AUROC & P-AUPR & AUPRO & Fit & I-AUROC & P-AUPR & AUPRO & Fit \\
\midrule
DINOv2 & k-means+PCA & \textbf{94.28}{\scriptsize$\pm$0.17} & \textbf{42.08}{\scriptsize$\pm$0.41} & \textbf{95.70}{\scriptsize$\pm$0.15} & 48\,s & \textbf{96.18}{\scriptsize$\pm$0.21} & \textbf{44.80}{\scriptsize$\pm$0.34} & \textbf{96.65}{\scriptsize$\pm$0.13} & 79\,s \\
 & EM, k-means start & 91.80{\scriptsize$\pm$0.41} & 38.80{\scriptsize$\pm$0.66} & 95.01{\scriptsize$\pm$0.14} & 25\,min & 94.88{\scriptsize$\pm$0.27} & 42.54{\scriptsize$\pm$0.52} & 96.38{\scriptsize$\pm$0.13} & 85\,min \\
\midrule
DINOv3 & k-means+PCA & \textbf{93.64}{\scriptsize$\pm$0.66} & \textbf{54.74}{\scriptsize$\pm$0.71} & \textbf{94.67}{\scriptsize$\pm$0.22} & 32\,s & \textbf{96.21}{\scriptsize$\pm$0.23} & \textbf{54.92}{\scriptsize$\pm$0.49} & \textbf{95.83}{\scriptsize$\pm$0.02} & 51\,s \\
 & EM, k-means start & 92.93{\scriptsize$\pm$0.54} & 50.70{\scriptsize$\pm$0.24} & 93.81{\scriptsize$\pm$0.28} & 8\,min & 95.58{\scriptsize$\pm$0.32} & 52.98{\scriptsize$\pm$0.52} & 95.52{\scriptsize$\pm$0.02} & 31\,min \\
\bottomrule
\end{tabular}
\end{adjustbox}
\end{table*}

\FloatBarrier
\newpage
\section{Layerwise Component Selection for DINOv3}
\label{sec:layer-selection}

The DINOv3 variant of LocalSubspaceAD builds local subspaces for each layer $l \in \mathcal{L} = \{16, 20, 24\}$ of DINOv3-L independently. Each layer describes a different semantic level between features. Therefore, we investigate two clustering strategies: i) shared clustering based on a single reference layer or ii) individual clustering per layer. \cref{tab:cluster_layer_visa} shows the performance of DINOv3-based LocalSubspaceAD with clustering performed in both ways. Shared clustering using reference layer 20 performs best. Both clustering with reference layer 24 and individual clustering decreases  performance over all metrics. Shared clustering using reference layer 16 yields better image-wise detection, but results in reduced P-AUPR. As best overall, shared clustering using reference layer 20 is used as the default selection.
\begin{table*}[t]
\centering
\caption{Shared clustering using different reference layers compared to individual clustering, per-shot, average over three seeds, VisA. Difference to LocalSubspaceAD with shared clustering layer 20 (the default), in points, mean $\pm$ std over the seeds of the paired difference.}
\label{tab:cluster_layer_visa}
\setlength{\tabcolsep}{3pt}
\begin{tabular}{l ccc ccc ccc}
\toprule
& \multicolumn{3}{c}{1-shot} & \multicolumn{3}{c}{2-shot} & \multicolumn{3}{c}{4-shot} \\
\cmidrule(lr){2-4}\cmidrule(lr){5-7}\cmidrule(lr){8-10}
Clustering from & I-AUROC & P-AUPR & AUPRO & I-AUROC & P-AUPR & AUPRO & I-AUROC & P-AUPR & AUPRO \\
\midrule
shared (layer 20) & 93.68{\tiny$\pm$0.68} & 55.04{\tiny$\pm$0.70} & 94.74{\tiny$\pm$0.20} & 95.04{\tiny$\pm$0.66} & 55.62{\tiny$\pm$0.33} & 95.45{\tiny$\pm$0.09} & 96.15{\tiny$\pm$0.20} & 54.81{\tiny$\pm$0.53} & 95.76{\tiny$\pm$0.01} \\
\midrule
shared (layer 16) & +0.14{\tiny$\pm$0.61} & -0.73{\tiny$\pm$0.70} & -0.09{\tiny$\pm$0.08} & +0.50{\tiny$\pm$0.17} & -1.05{\tiny$\pm$0.56} & +0.03{\tiny$\pm$0.27} & +0.23{\tiny$\pm$0.21} & -0.31{\tiny$\pm$0.29} & +0.11{\tiny$\pm$0.16} \\
shared (layer 24) & -2.12{\tiny$\pm$1.08} & -5.03{\tiny$\pm$0.22} & -3.26{\tiny$\pm$0.78} & -1.70{\tiny$\pm$0.08} & -3.87{\tiny$\pm$0.65} & -1.89{\tiny$\pm$0.29} & -1.68{\tiny$\pm$0.33} & -3.16{\tiny$\pm$0.21} & -0.86{\tiny$\pm$0.19} \\
individual & -0.38{\tiny$\pm$0.55} & -1.07{\tiny$\pm$0.48} & -0.40{\tiny$\pm$0.13} & -0.37{\tiny$\pm$0.19} & -1.34{\tiny$\pm$0.26} & -0.38{\tiny$\pm$0.20} & -0.22{\tiny$\pm$0.22} & -0.40{\tiny$\pm$0.44} & -0.09{\tiny$\pm$0.12} \\
\bottomrule
\end{tabular}
\end{table*}

\clearpage
\FloatBarrier
{
    \small
    \bibliographystyleS{ieeenat_fullname}
    \bibliographyS{_mppcaad}
}

\end{document}